\documentclass{article}

\usepackage{hello}

\usepackage{fancyhdr} 

\fancypagestyle{firstpage}{
    \fancyhf{}
    \lhead{\includegraphics[width=4cm]{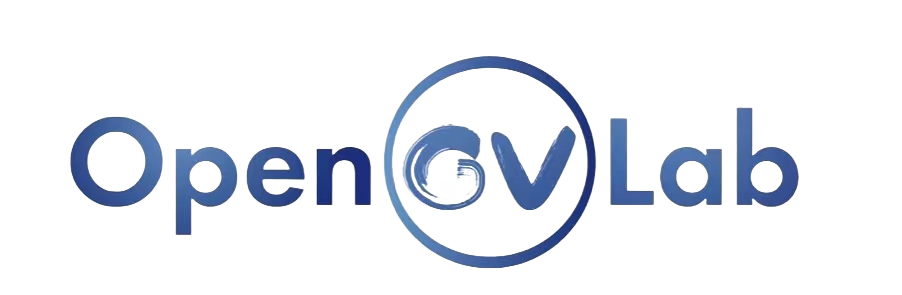}}
    \renewcommand{\headrulewidth}{0.4pt}
}
\usepackage[margin=1in]{geometry}

\usepackage{graphicx}
\usepackage{booktabs}
\usepackage{array}
\usepackage{amsmath}
\usepackage{amssymb}
\usepackage{amsfonts}
\usepackage{multirow}
\usepackage{verbatim}
\usepackage{caption}
\usepackage{longtable}
\usepackage{supertabular}
\usepackage{CJKutf8}
\usepackage[utf8]{inputenc} 
\usepackage[T1]{fontenc}
\usepackage[french,vietnamese,mongolian,greek,english]{babel}

\usepackage{enumitem}
\usepackage{tablefootnote}
\usepackage[numbers]{natbib}
\usepackage{colortbl}
\usepackage{xspace}
\usepackage{textcomp}
\usepackage{makecell}
\usepackage{lscape} 
\usepackage{siunitx}
\usepackage{listings}
\usepackage{color}
\usepackage{xcolor}

\definecolor{dt}{gray}{0.7}
\definecolor{cvprblue}{rgb}{0.21,0.49,0.74}
\definecolor{prompt}{rgb}{0.21,0.49,0.74}
\usepackage{pifont}       
\usepackage{bbding}       
\usepackage{fontawesome}

\usepackage{scrextend}

\usepackage{tgpagella}
\usepackage{latexsym}
\usepackage[T1]{fontenc}
\usepackage[utf8]{inputenc}
\usepackage{microtype}
\definecolor{mydarkblue}{rgb}{0,0.08,0.45}
\definecolor{citecolor}{HTML}{0071BC}
\usepackage[backref=page,pagebackref=true,colorlinks,citecolor=citecolor,urlcolor=magenta,linkcolor=magenta]{hyperref}
\usepackage{url}            
\usepackage{nicefrac}       
\usepackage{changepage}
\usepackage{xargs}          
\usepackage{wrapfig,lipsum,booktabs}
\usepackage{longtable}
\usepackage{subcaption}
\usepackage{endnotes}

\usepackage{pgfplots}
\usetikzlibrary{pgfplots.groupplots}
\pgfplotsset{compat=1.3}
\usepackage{tikz}
\usetikzlibrary{patterns}

\usepackage[most]{tcolorbox}

\usepackage[capitalize,noabbrev]{cleveref}
\crefname{section}{Section}{\S\S}
\Crefname{section}{Section}{\S\S}
\crefname{table}{Table}{Tables}
\crefname{figure}{Figure}{Figures}
\crefname{algorithm}{Algorithm}{}
\crefname{equation}{eq.}{}
\crefname{appendix}{Appendix}{}
\crefformat{section}{Section #2#1#3}
\usepackage{multicol}
\usepackage{tcolorbox}

\usepackage{titlesec}
\titleformat*{\section}{\large\bfseries}

\definecolor{battleshipgrey}{rgb}{0.3, 0.3, 0.3}
\definecolor{brilliantrose}{rgb}{1.0, 0.33, 0.64}
\definecolor{americanrose}{rgb}{1.0, 0.01, 0.24}
\definecolor{jweigreen}{rgb}{0,0.45,0.24}
\definecolor{bluegray}{rgb}{0.1, 0.1, 0.4}
\definecolor{ao(english)}{rgb}{0.0, 0.5, 0.0}
\definecolor{blanchedalmond}{rgb}{1.0, 0.92, 0.8}
\definecolor{atomictangerine}{rgb}{1.0, 0.6, 0.4}
\definecolor{chocolate(web)}{rgb}{0.82, 0.41, 0.12}
\definecolor{bananayellow}{rgb}{1.0, 0.88, 0.21}
\definecolor{goldenbrown}{rgb}{0.6, 0.4, 0.08}
\definecolor{aliceblue}{rgb}{0.94, 0.97, 1.0}
\definecolor{beige}{rgb}{0.96, 0.96, 0.86}
\definecolor{babyblue}{rgb}{0.54, 0.81, 0.94}
\definecolor{camel}{rgb}{0.76, 0.6, 0.42}
\definecolor{cinnamon}{rgb}{0.82, 0.41, 0.12}
\definecolor{frenchblue}{rgb}{0.0, 0.45, 0.73}
\definecolor{classicrose}{rgb}{0.98, 0.8, 0.91}
\definecolor{frenchrose}{rgb}{0.96, 0.29, 0.54}
\definecolor{frenchlilac}{rgb}{0.53, 0.38, 0.56}
\definecolor{frenchbeige}{rgb}{0.65, 0.48, 0.36}
\definecolor{verylightgreen}{RGB}{240, 255, 235}
\definecolor{verylightred}{RGB}{255, 235, 235}
\definecolor{verylightyellow}{RGB}{255, 254, 235}
\definecolor{dt}{gray}{0.7}

\definecolor{forestgreen}{HTML}{2e7d43}
\definecolor{color1}{HTML}{FF9999}
\definecolor{color2}{HTML}{FF6666}
\definecolor{color3}{HTML}{FF3333}
\definecolor{color4}{HTML}{E60000}
\definecolor{color5}{HTML}{B30000}
\definecolor{color6}{HTML}{8CD98C}
\definecolor{color7}{HTML}{53c653}
\definecolor{color8}{HTML}{39ac39}
\definecolor{color9}{HTML}{2d862d}
\definecolor{color10}{HTML}{206020}
\definecolor{color11}{HTML}{cca300}

\usepackage{minitoc}
\usepackage{float}

\newlength\savewidth

\makeatletter
\def\blfootnote{\xdef\@thefnmark{}\@footnotetext}
\makeatother

\usepackage{amsmath,amsfonts,bm}

\def\eqref#1{equation~\ref{#1}}

\def\1{\bm{1}}

\DeclareMathAlphabet{\mathsfit}{\encodingdefault}{\sfdefault}{m}{sl}
\SetMathAlphabet{\mathsfit}{bold}{\encodingdefault}{\sfdefault}{bx}{n}

\title{
\textbf{
Mono-InternVL: Pushing the Boundaries of  Monolithic Multimodal Large Language Models with Endogenous Visual Pre-training}
}

\author{
    \normalsize{}
    \textbf{Gen Luo$^{1*}$,~Xue Yang$^{1*}$,~Wenhan Dou$^{2*}$,~Zhaokai Wang$^{3,1*}$}\\
    \normalsize{}
    ~\textbf{~Jiawen Liu$^4$, ~Jifeng Dai$^{2,1}$, ~Yu Qiao$^1$, ~Xizhou Zhu$^{2,1\dag}$} \\ %
    \normalsize{}
    ~$^1$OpenGVLab, Shanghai AI Laboratory \quad $^2$Tsinghua University 
    \\ ~~\normalsize{}$^3$Shanghai Jiao Tong University \quad $^4$Johns Hopkins University  \\
    [3mm]
    \normalsize{}
    Project Page: 
    \href{https://internvl.github.io/blog/2024-10-10-Mono-InternVL/}{Mono-InternVL} \\
    \normalsize{}
    Code: 
    {\url{https://github.com/OpenGVLab/Mono-InternVL}}
}

\date{}

\begin{document}

\doparttoc 
\faketableofcontents 

\maketitle
\blfootnote{\noindent$^{*}$Equal contribution. $^{\dag}$Corresponding author.}

\begin{abstract}
\noindent
In this paper, we focus on  monolithic Multimodal Large Language Models (MLLMs) that integrate visual encoding and language decoding into a single LLM.  In particular, we identify that existing pre-training strategies for monolithic MLLMs often suffer from unstable optimization or catastrophic forgetting.  To address this issue, our core idea is to embed a new visual parameter space into a pre-trained LLM, thereby stably learning visual knowledge from noisy data while freezing the LLM.  Based on this principle, we   present Mono-InternVL,  a novel monolithic MLLM  that  seamlessly integrates a set of  visual experts via  a multimodal mixture-of-experts structure.  Moreover, we propose an innovative pre-training strategy to maximize the visual capability of Mono-InternVL, namely Endogenous Visual Pre-training (EViP).  In particular, EViP is designed as a progressive learning process for  visual experts,   which aims to  fully exploit the visual knowledge  from noisy data to high-quality data.   To validate our approach, we conduct extensive experiments on 16 benchmarks.  Experimental results confirm the superior performance of Mono-InternVL than existing monolithic MLLMs on 13 of 16 multimodal benchmarks, \emph{e.g.,} +80 points over Emu3 on OCRBench.  Compared to the modular baseline, \emph{i.e.,} InternVL-1.5, Mono-InternVL  still retains comparable multimodal performance while reducing up to   67\% first token latency.  Code and model are released at \url{https://github.com/OpenGVLab/Mono-InternVL}.

\end{abstract}

\thispagestyle{firstpage} 

\section{Introduction}
\begin{figure*}[t]
    \centering
    \includegraphics[width=0.9\linewidth]{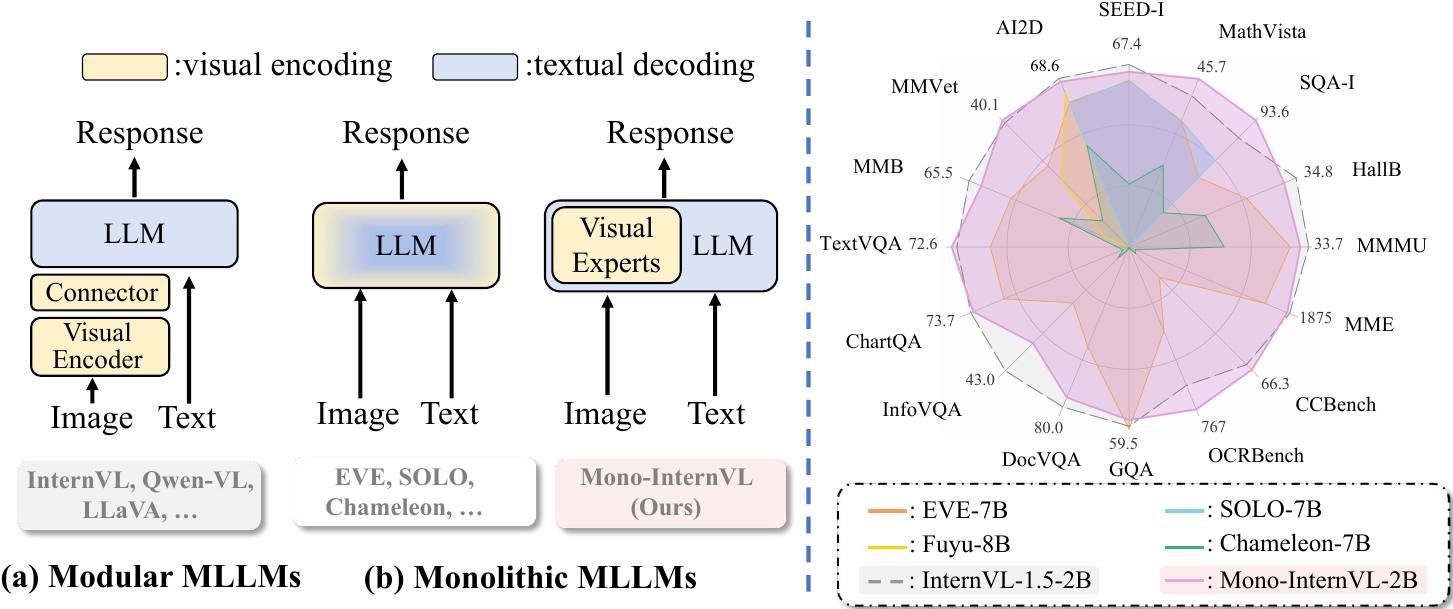}
    {
    \vspace{-0.5em}
    \caption{\textbf{Comparison of  Mono-InternVL  and  existing MLLMs.} 
     Compared with modular MLLMs, Mono-InternVL embeds visual experts into the pre-trained LLM and integrates visual encoding and language decoding into a single LLM. Through endogenous visual pre-training, Mono-InternVL significantly pushes the performance boundaries of monolithic MLLMs.
    \label{fig:fig1}
    }
    \vspace{-0.5em}
    }
\end{figure*}

In recent years, the rapid development of Large Language Models (LLMs)~\citep{VLM:GPT-4, TransF:Qwen, cai2024internlm2} has spurred increasing efforts to extend their multimodal capabilities~\citep{ VLM:InternVL, VLM:LLaVA}.  As shown in Fig.~\ref{fig:fig1} (a), most existing Multimodal Large Language Models (MLLMs) adopt a modular architecture, where visual encoding and language decoding are processed separately. This is typically achieved by combining a pre-trained visual encoder like CLIP-ViT~\citep{VLP:CLIP} with an LLM~\citep{VLM:LLaVA,   VLM:InternVL-1.5, VLP:BLIPv2}. Recent research has also started exploring monolithic MLLMs~\citep{VLM:Fuyu-8b, diao2024EVE, solo}, as shown in Fig.~\ref{fig:fig1} (b), which integrate visual perception and multimodal understanding directly into a single LLM.  While modular MLLMs achieve superior performance, monolithic MLLMs show better potential in terms of design simplicity and deployment efficiency~\citep{diao2024EVE,solo}.

Despite recent progress, training a monolithic MLLM with promising performance still remains challenging. In particular, monolithic MLLMs struggle to replicate the successful usage of pre-trained visual encoders in modular MLLMs for visual perception. Therefore, researchers often rely on additional pre-training to compensate for the shortcomings in visual perception in monolithic MLLMs~\citep{team2024chameleon,diao2024EVE}. 
A straightforward approach is the \textit{native pre-training}~\citep{team2024chameleon}, which pre-trains a monolithic MLLM from scratch using a mixture of text-only and multimodal data. However, it requires prohibitively high training costs and often suffers from challenges of unstable optimization~\citep{team2024chameleon}. 
Another common solution is the \textit{continuous pre-training}~\citep{diao2024EVE}, which  extends the pre-trained LLM  to multimodality via additional visual pre-training.  Benefiting from the knowledge in the pre-trained LLM, the cost of continuous pre-training becomes much more affordable. Nevertheless, due to the catastrophic forgetting issue~\citep{zhai2023investigating}, the pre-trained language knowledge is inevitably undermined during continuous pre-training, thereby weakening the multimodal capabilities.

In this paper, we aim to address the forgetting issue of continuous pre-training  through \textit{delta tuning}~\citep{ding2022delta}.
In particular, we argue that such issue arises from the shared architecture for joint vision and language modeling, where optimizations for vision can negatively impact language capabilities.
Therefore, it is a natural thought to introduce an independent visual parameter set into the pre-trained LLM, which allows formulating visual pre-training with partial parameter tuning. 
This can help retain the language knowledge by freezing the entire LLM during continuous pre-training, while improving visual learning.
This principle is also aligned with previous endeavors in modular MLLMs, \emph{e.g.,} QwenVL~\citep{bai2023qwenvl} and InternVL-1.5~\citep{VLM:InternVL-1.5},  where the visual parameters are placed outside the LLM. 

\begin{figure*}[t!]
    \centering
    \includegraphics[width=0.85\linewidth]{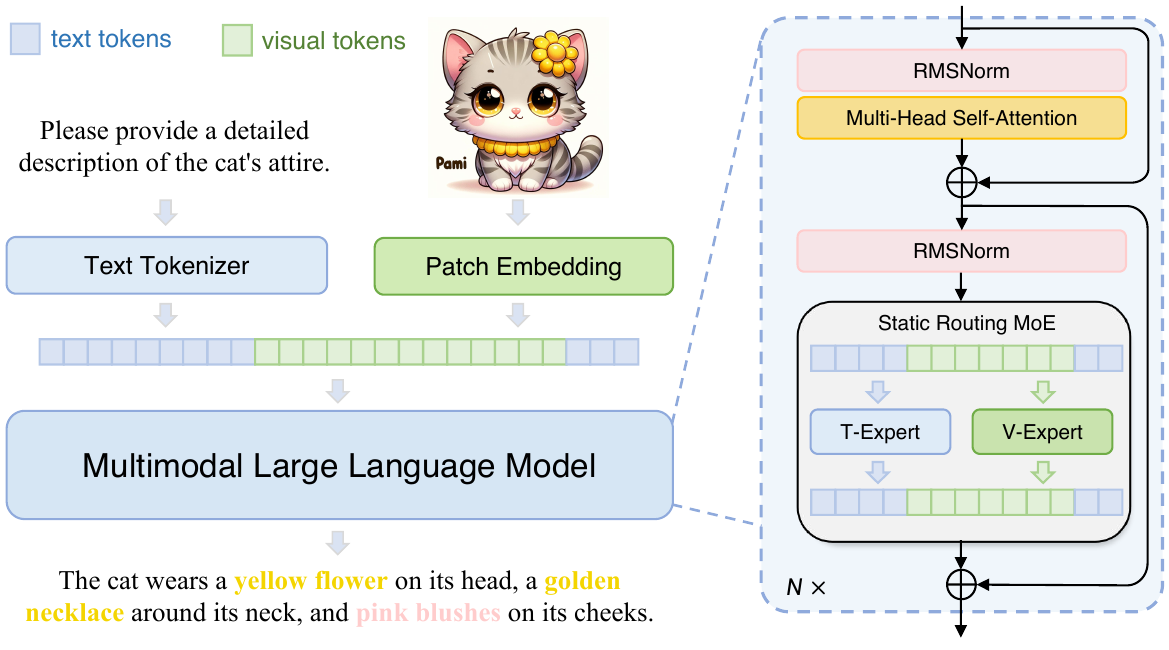}
    {
    \vspace{-.2em}
    \caption{\textbf{Monolithic architecture of Mono-InternVL.}    Mono-InternVL is designed as a multimodal MoE structure, where  visual  and textual tokens are processed by the corresponding experts.  This design  facilitates the visual pre-training while retaining the model efficiency.}
    \label{fig:framework}
    }
\vspace{-8pt}
\end{figure*}

Based on the above principle, we propose a novel monolithic MLLM, namely Mono-InternVL. As shown in Fig.~\ref{fig:framework}, the visual parameters in Mono-InternVL are designed as a set of expert networks that are seamlessly integrated into the LLM via the mixture-of-experts mechanism. To better capture the visual knowledge, these experts are initialized from the Feed-Forward Networks (FFNs) in the pre-trained LLM. Based on this architecture, we present an innovative \textit{Endogenous Visual Pre-training} (EViP) method. Specifically, EViP is formulated as a progressive learning process of three stages: 1) concept learning to grasp basic visual concepts, 2) semantic learning to capture high-level semantics, \emph{e.g.,} world knowledge, and 3) alignment learning to align knowledge with downstream tasks.   Benefiting from the architecture and the pre-training strategy, the visual scalability of Mono-InternVL is fully unleashed,  where the downstream performance consistently improves as the scale of the pre-training data increases.  After visual pre-training, Mono-InternVL accommodates complex multimodal tasks via supervised instruction tuning.  

To validate our method, we develop Mono-InternVL using the pre-trained LLM InternLM2-1.8B~\citep{cai2024internlm2}, and conduct extensive experiments on 16 multimodal benchmarks.  Experimental results  demonstrate the significant performance improvements of  Mono-InternVL against previous monolithic MLLMs. For instance, Mono-InternVL with 1.8 billion activated parameters  can outperform existing monolithic MLLMs with 8 billion parameters, \emph{e.g.,} +2.5\%  over Emu3~\cite{emu3} on average.   Compared to the modular baseline, \emph{i.e.,} InternVL-1.5~\citep{VLM:InternVL-1.5}, Mono-InternVL shows comparable performance on 16 multimodal benchmarks while  reducing  first token latency by  67\%.   In conclusion, our contributions can be summarized in threefold:
\begin{itemize}[leftmargin=*]
    \item We present Mono-InternVL,  a novel monolithic MLLM that seamlessly integrates a set of  visual experts via  a multimodal mixture-of-experts architecture.  This architecture effectively extends the pre-trained LLM to a monolithic MLLM  while retaining the pre-trained knowledge.
    \item We propose a novel visual pre-training approach for Mono-InternVL called endogenous visual pre-training (EViP).  EViP adopts a progressive learning strategy to encourage visual experts to continuously grasp visual knowledge from noisy data to high-quality data.
    \item  Mono-InternVL demonstrates for the first time that  monolithic MLLMs can reach the  comparable  performance with  leading modular MLLMs, opening new avenues for designing future MLLMs.   
\end{itemize}

\section{Related Work}
\textbf{Modular multimodal large language models.~}
Recent progress in large language models (LLMs) has catalyzed the integration of vision and language modalities, giving rise to multimodal large language models (MLLMs). Both commercial entities, such as GPT-4V~\citep{VLM:GPT-4v} and Gemini series~\citep{VLM:Gemini}, and other open-source initiatives, \emph{e.g.} BLIP series~\citep{VLP:BLIP,VLP:BLIPv2,VLM:InstructBLIP}, LLaVA series~\citep{VLM:LLaVA,VLM:LLaVA-1.5,VLM:LLaVA-1.6}, InternVL~\citep{VLM:InternVL,VLM:InternVL-1.5}, have pursued this integration, often linking LLMs~\citep{TransF:LLaMA,TransF:LLaMA2,TransF:InternLM,cai2024internlm2} with large vision models (LVMs)~\citep{VLP:CLIP,TransF:ViT,TransF:ViT-22B,VLM:InternVL} via intermediate layers. 
However, such encoder-based vision-language models (modular MLLMs) encounter challenges like limitations in visual processing due to pre-trained  encoders, deployment inefficiencies, and complexities in balancing the capacities of LLMs and LVMs, as  pointed out by~\citep{diao2024EVE}.

\textbf{Monolithic multimodal large language models.~}
The issues related to modular MLLMs have steered research toward encoder-free architectures, also known as monolithic MLLMs, which can be summarized into two categories. The first category obtains continuous visual tokens through a lightweight structure before feeding into MLLMs. For instance, Fuyu-8B~\citep{VLM:Fuyu-8b} processes images directly through a simple linear projection, adeptly handling high-resolution images without using a dedicated visual encoder. EVE-7B~\citep{diao2024EVE} prioritizes vision-language pre-alignment from an LLM-centric perspective and enhances image recognition through visual distillation. SOLO~\citep{solo} introduces an open-source training recipe for developing monolithic MLLMs. In contrast, the second category introduces VQ tokenizer-based models to generate discrete visual tokens for image generation, with representative works like Chameleon~\citep{team2024chameleon}, Show-o~\citep{xie2024show}, Transfusion~\citep{zhou2024transfusion} and Emu3~\citep{emu3}.

\textbf{Multimodal mixture-of-experts.~}
VLMo~\citep{bao2022vlmo} and BEiT-3~\citep{VLP:BEiTv3} employ a pool of modality experts to replace the feed-forward network in standard Transformer to capture modality-specific information by switching to different modality experts, and use the shared self-attention across modalities to align visual and linguistic information.
VL-MoE~\citep{shen2023scaling} introduces mixture-of-experts (MoE)~\citep{yuksel2012twenty} based on the above works for better training and deploying.
MoMa~\citep{lin2024moma} also uses multimodal mixture-of-experts for pre-training of MLLMs~\citep{team2024chameleon} and cooperates with sparse components, \emph{e.g.} MoE and mixture-of-depths (MoD)~\citep{raposo2024mixture} to improve the efficiency of pre-training from scratch with trillions of mixed-modal tokens. Inspired by the above literature, we introduce multimodal mixture-of-experts (\emph{i.e.,} a visual expert and a language expert) for pre-training monolithic MLLMs and use a novel progressive learning strategy, named endogenous visual pre-training (EViP), to address the unique challenges faced by monolithic MLLMs.

\section{Mono-InternVL} 
\subsection{The Monolithic Architecture}
As shown in Fig.~\ref{fig:framework}, we first outline the architecture of Mono-InternVL, which consists of tokenizers and a multimodal mixture-of-experts structure.

\textbf{Visual and textual embeddings.} Compared to modular MLLMs,  Mono-InternVL directly patchifies images to input visual sequences using a lightweight module.  Specifically, given the input image $I\in \mathbb{R}^{H\times W \times 3}$, the input visual embedding $x_v \in \mathbb{R}^{(h \times w) \times d}$ is obtained by
\begin{equation}
\begin{aligned}
x_v=\text{MLP}(\text{PatchEmbed}(I)+ \text{PE}).
\end{aligned}
\end{equation}
Here, $\text{PatchEmbed}(\cdot)$  denotes a patch embedding layer with a stride of 28,  meaning each visual token represents a  $28\times 28$ image patch.   $\text{PE} \in \mathbb{R}^{(h \times w) \times d}$ is the learnable positional embedding as similar to InternVL-1.5~\citep{VLM:InternVL-1.5}. Besides, we also add  an additional thumbnail to provide global visual information. 
After that, an MLP layer, \emph{i.e.,} $\text{MLP}(\cdot)$, is used to project visual patches into the $d$-dimensional embedding space of the LLM.  This simple visual tokenizer allows Mono-InternVL to process  arbitrary-resolution images with up to 8 millions of pixels, \emph{i.e.,} $10,240$ image patches,  which  can cover most high-resolution scenarios. 

In  Mono-InternVL, the textual tokenizer remains the same as the original one in the LLM. Given  the input text $T \in \mathbb{Z}^n$,   we obtain textual embedding $x_t \in \mathbb{R}^{n\times d}$ by  
\begin{equation}
\begin{aligned}
x_t=\text{Tokenizer}(T).
\end{aligned}
\end{equation}
Afterwards, the multimodal embedding is constructed as the concatenation of visual and textual embeddings, denoted as $x_m \in \mathbb{R}^{n' \times d}$.

\textbf{Multimodal mixture-of-experts structure.}
The key principle of Mono-InternVL is to embed visual experts into a pre-trained LLM. In this case, Mono-InternVL can not only facilitate  the visual pre-training with the pre-trained LLM knowledge, but also significantly mitigates the catastrophic forgetting issue during pre-training.
Specifically, given  the multimodal input $x_m \in \mathbb{R}^{n' \times d }$,  a decoder-only LLM  with a set of visual experts is used to generate the  textual tokens step by step, which can be formulated by 
\begin{equation}
    \begin{aligned}
        p_s= \mathcal{F_\text{llm}}(y_{s}| x_m, y_{0:s-1};\theta,\theta_v).
    \end{aligned} 
    \label{eq_arch}
\end{equation}
Here, $y \in \mathbb{R}^{S}$  and $S$ denote the word length   and its length, respectively.  $p_s\in \mathbb{R}^{m}$ is the  next-token probability and $m$ is the size of word vocabulary. $\mathcal{F}_\text{llm}$  and $\theta$ denote the LLM  and its  pre-trained parameters, respectively. $\theta_v$ is the parameters of patch embedding layer and  visual experts. 

As shown in Fig.~\ref{fig:framework},  $\mathcal{F}_\text{llm}$ is designed as a multimodal mixture-of-experts structure.  In particular, we adopt the static routing strategy that assigns  visual and textual experts to the corresponding tokens. Therefore, the \textit{l}-th LLM layer  can be defined by
\begin{equation}
    \begin{aligned}
        x_m^{l'}&=x_m^{l-1}+\text{MHA}(\text{RMSNorm}(x_m^{l-1})),\\
        x_m^{l}&=x_m^{l'}+\text{MMoE}(\text{RMSNorm}(x_m^{l'})).
    \end{aligned}
    \label{mmoe_layer}
\end{equation}
Here, $\text{MHA}(\cdot)$ and $\text{RMSNorm}(\cdot)$ denote the multi-head attention~\citep{TransF:Transformer} and the layer normalization~\citep{zhang2019root}, respectively.  $\text{MMoE}(\cdot)$ is the proposed multimodal mixture-of-experts, formulated by
\begin{equation}
\begin{aligned}
    \text{MMoE}(x)=
\begin{cases} 
\text{FFN}_v(x) \quad \text{if } x \in x_v, \\
\text{FFN}_t(x) \quad \text{if } x \in x_t.
\end{cases}
\end{aligned}
\label{mmoe_module}
\end{equation}
 Here, $x \in \mathbb{R}^d$ is the element of $x_m$. $\text{FFN}_v$ and $\text{FFN}_t$ denote the visual and textual experts, respectively.   In practice, $\text{FFN}_v$ is initialized from the  $\text{FFN}_t$ to leverage the pre-trained knowledge.  

As defined in Eq.~\ref{mmoe_layer} and \ref{mmoe_module}, the MMoE structure has two distinct advantages over the existing monolithic MLLM. Firstly, the visual learning of Mono-InternVL can greatly benefit from the pre-trained language knowledge, while the language ability can still be preserved by freezing $\text{FFN}_t$. Secondly, the MMoE structure significantly enhances the model's capacity for vision-and-language modeling, while the additional inference cost is almost negligible due to the MoE mechanism.

\begin{table*}[t]
\renewcommand{\arraystretch}{1.1}
\centering
\normalsize
\resizebox{0.99\linewidth}{!}{
\begin{tabular}{ccl}
\toprule
 \textbf{Stage} & \textbf{\#Samples} & \textbf{Datasets} \\
 \midrule
S1.1 &  922M  & Laion-EN (en)~\cite{Datasets:Laion-5b}, COYO (en)~\cite{kakaobrain2022coyo-700m}\\
  \midrule
  S1.2 &  258M  & Laion-EN (en)~\cite{Datasets:Laion-5b}, COYO (en)~\cite{kakaobrain2022coyo-700m}, SAM (en)~\cite{TransF:SAM}        \\
  \midrule
  &   & \textbf{Captioning:} Laion-EN (en)~\cite{Datasets:Laion-5b}, Laion-ZH (zh)~\cite{Datasets:Laion-5b}, COYO (zh)~\cite{kakaobrain2022coyo-700m},GRIT (zh)~\cite{peng2023kosmos2}, COCO (en)~\cite{chen2015cococaption},  TextCaps (en)~\cite{Datasets:TextCaps} \\
  &   & \textbf{Detection:} Objects365 (en\&zh)~\cite{shao2019objects365}, GRIT (en\&zh)~\cite{peng2023kosmos2},  All-Seeing (en\&zh)~\cite{wang2023allseeing}    \\
  &   & \textbf{OCR (large):} Wukong-OCR (zh)~\cite{gu2022wukong}, LaionCOCO-OCR (en)~\cite{Datasets:LAION-COCO},  Common Crawl PDF (en\&zh) \\
  &   & \textbf{OCR (small):} MMC-Inst (en)~\cite{liu2023mmcinst}, LSVT (zh)~\cite{sun2019lsvt}, ST-VQA (en)~\cite{Datasets:STVQA}, RCTW-17 (zh)~\cite{shi2017rctw17},ReCTs (zh)~\cite{zhang2019rects},   ArT (en\&zh)~\cite{chng2019art},   \\
  &   &   SynthDoG (en\&zh)~\cite{kim2022synthdog}, COCO-Text (en)~\cite{veit2016cocotext}, ChartQA (en)~\cite{Datasets:ChartQA}, CTW (zh)~\cite{yuan2019ctw}, DocVQA (en)~\cite{Datasets:DocVQA}, TextOCR (en)~\cite{singh2021textocr}, \\
  \multirow{-6}{*}{S1.3} &  \multirow{-6}{*}{143M} &  PlotQA (en)~\cite{methani2020plotqa}, InfoVQA (en)~\cite{mathew2022infographicvqa}    \\
  \bottomrule
\end{tabular}
}
\caption{\textbf{Summary of datasets used in the endogenous visual pre-training.}  In S1.2,  caption for each image is synthetically produced by the pre-trained InternVL2-8B~\cite{VLM:InternVL-1.5}.
}
\label{tab:datasets_pretrain}
\vspace{-8pt}
\end{table*}

\begin{figure*}[t!]
    \centering
    \includegraphics[width=0.99\linewidth]{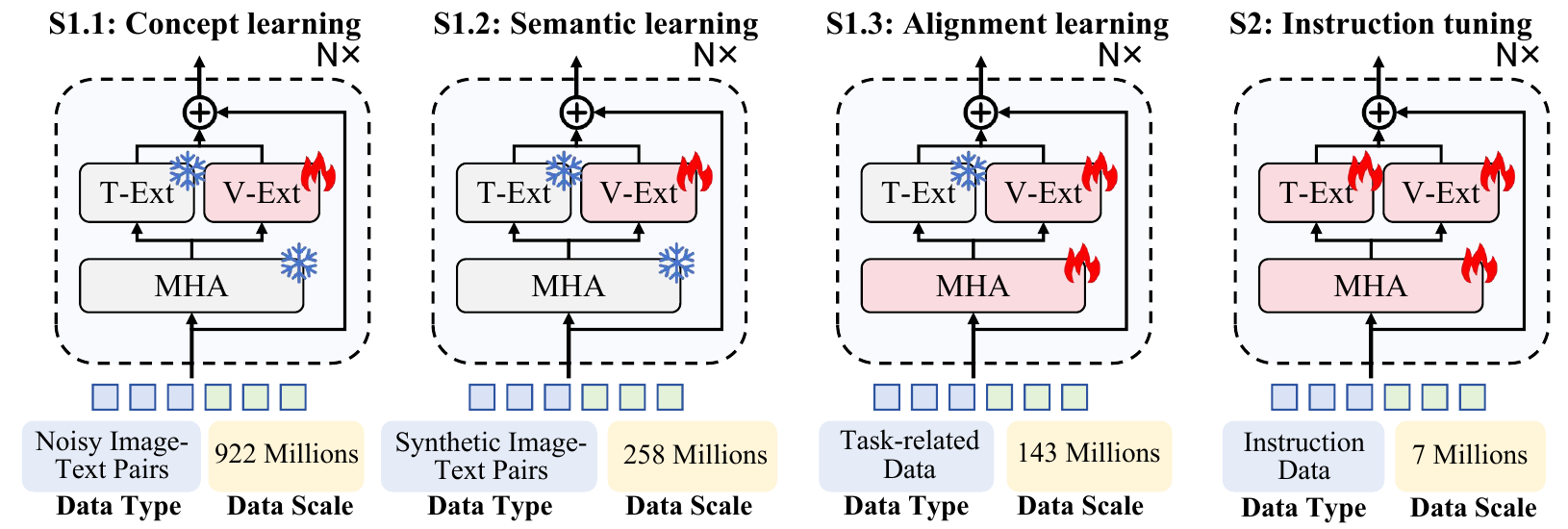}
    {
    \vspace{-.5em}
    \caption{\textbf{The training recipe of Mono-InternVL.}  In the first stage, Mono-Internvl is progressively pre-trained on massive data via three sub-stages (S1.1, S1.2, S1.3), where most parameters of LLM are frozen to preserve the pre-trained knowledge. In the second stage (S2), the entire model is optimized to accommodate various instructions. }
    \label{fig:fig2}
    }
\vspace{-8pt}
\end{figure*}

\subsection{Endogenous Visual Pre-training}

Endogenous Visual Pre-training (EViP) aims to maximize the benefits of Mono-InternVL from visual experts through pre-training on massive noisy and synthetic data.  Unlike existing methods~\citep{diao2024EVE,team2024chameleon}, we formulate EViP from the perspective of delta tuning~\citep{ding2022delta}, in which most of the LLM parameters are frozen to preserve its pre-trained knowledge.  Therefore, the objective of EViP can be defined by 
\begin{equation}
    \begin{aligned}
        \arg \min_{\Delta \theta} \mathcal{L}(\mathcal{F}_{\text{llm}}(x_m; \theta, \theta_v), \hat{y}),
    \end{aligned}
    \label{delta_tuning}
\end{equation}
  where $\mathcal{L}(\cdot)$ and $\hat{y}$ denote the auto-regressive loss and the ground-truth, respectively. As shown in Fig.~\ref{fig:fig2}, $\Delta \theta$ denotes parameters of patch embedding and visual experts in the concept and semantic learning, \emph{i.e.,} $\theta_v$, while in the alignment learning stage  $\Delta \theta$ also includes the parameters of multi-head attentions. 
Based on Eq.~\ref{delta_tuning}, EViP is designed as a progressive learning process.  As shown in Fig.~\ref{fig:fig2} and Tab.~\ref{tab:datasets_pretrain}, EViP consists of  three sub-stages, namely concept learning (S1.1), semantic learning (S1.2) and alignment learning (S1.3).  For different sub-stages, we use carefully partitioned data to achieve coarse-to-fine visual learning. 

\textbf{Concept learning.}  Concept learning aims to encourage the model to learn fundamental visual concepts, such as object categories or basic shapes.  Therefore, we first pre-train Mono-InternVL with  about 922 million noisy samples, which are sampled from Laion-2B~\citep{Datasets:Laion-5b} and Coyo-700M~\citep{kakaobrain2022coyo-700m}.   In this sub-stage, Mono-InternVL employs a simple prompt to perform generative learning, \emph{i.e.,} ``provide a one-sentence caption for the image''. Meanwhile, we constrain the maximum   number of image patches of the  visual tokenizer to 1,280  for training efficiency. To preserve the language capabilities while enabling visual specialization, the entire LLM is kept frozen during concept learning, and only  the patch embedding and visual experts  are optimized. 

\textbf{Semantic learning.} After concept learning, Mono-InternVL is able to understand  basic concepts in the image, but organizing this information to produce reasonable descriptions remains challenging.  To achieve a higher-level visual understanding,  we utilize the pre-trained InternVL2-8B~\citep{VLM:InternVL-1.5} to produce  short captions for 258 million images.  Compared to the original noisy captions,  synthetic captions depict complex visual knowledge, such as relationship and world knowledge, \textit{etc.}, while containing less noisy information unrelated to the image, \emph{e.g.,} time of shooting or the photographer. In this sub-stage, we adopt the same optimization strategy as concept learning, except that the maximum number of image patches is increased to 1,792.
 
\textbf{Alignment learning.}   To meet the visual requirements of downstream tasks, we further perform alignment learning on Mono-InternVL. As shown in Tab.~\ref{tab:datasets_pretrain}, our alignment data is sampled from the pre-training data of InternVL-1.5~\citep{VLM:InternVL-1.5}, including 143 million  samples of image captioning, detection and optical character recognition (OCR).  In particular, captioning data, detection data and OCR data account for about 53.9\%, 5.2\% and 40.9\% of the total, respectively.  In this sub-stage, we utilize the task-specific prompts from InternVL-1.5 for the generative learning, and increase the maximum number of image patches to 3,328. Compared to previous sub-stages,   the multi-head attention layers are additionally optimized  to achieve better vision-language alignment.

\subsection{Instruction Tuning}
In this stage, we follow InternVL~\citep{VLM:InternVL-1.5} to use around 5 million bilingual instructions for supervised learning, covering various tasks such as visual question answering, multimodal dialogue, mathematics, knowledge, \textit{etc}.  
In this stage, the entire models are optimized and the maximum number of image patches is increased to 6,400 to accommodate high-resolution images.  Details of instruction data can be found in Appendix \S\ref{sec:supp_full_data_detail}.

\section{Experiments}
\begin{table*}[t!]
\centering
\setlength\tabcolsep{1.5pt}
\renewcommand{\arraystretch}{1.15}
\resizebox{\linewidth}{!}{
\begin{tabular}{lc|ccccccccc|c|ccccccc|c}
\toprule
\textbf{Model} & \textbf{\#A-Param} & \textbf{MMB} & \textbf{MMVet} & \textbf{MMMU} & \textbf{MME} & \textbf{MathVista} & \textbf{SEED-I} & \textbf{OCRBench} & \textbf{HallB} & \textbf{CCB} & \textbf{Avg$_{\text{MM}}$ }& \textbf{TextVQA} & \textbf{SQA-I} & \textbf{GQA} & \textbf{DocVQA} & \textbf{AI2D} & \textbf{ChartQA} & \textbf{InfoVQA} & \textbf{Avg$_{\text{QA}}$} \\
\midrule
\rowcolor{gray!15} \multicolumn{20}{l}{$\blacktriangledown$ \emph{Modular MLLMs:}}  \\
MobileVLM-V2-3B~\cite{chu2024mobilevlm}   & 3.0B & 63.2 & $-$ & $-$ & $-$ & $-$ & $-$ & $-$ & $-$ & $-$ & $-$ & 57.5 & 70.0 & 66.1 & $-$ & $-$ & $-$ & $-$ & $-$ \\
Mini-Gemini-2B~\cite{VLM:MiniGemini}    & 3.5B & 59.8 & 31.1 & 31.7 & 1653 & 29.4 & $-$ & $-$ & $-$ & $-$ & $-$ & 56.2 & $-$ & $-$ & 34.2 & $-$ & $-$ & $-$ & $-$ \\
MM1-3B-MoE-Chat~\cite{VLM:MM1}   & 3.5B & 70.8 & 42.2 & 38.6 & 1772 & 32.6 & 69.4 & $-$ & $-$ & $-$ & $-$ & 72.9 & 76.1 & $-$ & $-$ & $-$ & $-$ & $-$ & $-$ \\
DeepSeek-VL-1.3B~\cite{lu2024deepseekvl}  & 2.0B & 64.6 & 34.8 & 32.2 & 1532 & 31.1 & 66.7 & 409 & 27.6 & 37.6 & 43.4 & 57.8 & $-$ & $-$ & $-$ & 51.5 & $-$ & $-$ & $-$ \\
PaliGemma-3B~\cite{beyer2024paligemma}      & 2.9B & 71.0 & 33.1 & 34.9 & 1686 & 28.7 & 69.6 & 614 & 32.2 & 29.6 & 46.7 & 68.1 & $-$ & $-$ & $-$ & 68.3 & $-$ & $-$ & $-$ \\
MiniCPM-V-2~\cite{yao2024minicpm}       & 2.8B & 69.1 & 41.0 & 38.2 & 1809 & 38.7 & 67.1 & 605 & 36.1 & 45.3 & 51.2 & 74.1 & $-$ & $-$ & 71.9 & 62.9 & $-$ & $-$ & $-$ \\
$^\dagger$InternVL-1.5-2B~\cite{VLM:InternVL-1.5}   & 2.2B & 70.9 & 39.3 & 34.6 & 1902 & 41.1 & 69.8 & 654 & 37.5 & 63.5 & 54.4 & 70.5 & 84.9 & 61.6 & 85.0 & 69.8 & 74.8 & 55.4 & 71.7 \\
Qwen2VL-2B~\cite{Qwen2vl} & 2.1B & 74.9 & 49.5 &41.1 & 1872 & 43.0& -& 809 & 41.7& -& -& 79.7 &-& -& 90.1& 74.7& 73.5 & 65.5 &-  \\
\midrule
\rowcolor{gray!15} \multicolumn{20}{l}{$\blacktriangledown$ \emph{Monolithic MLLMs:}}   \\
Fuyu-8B (HD)~\cite{VLM:Fuyu-8b} & 8B & 10.7 & 21.4 & $-$ & $-$ & $-$ & $-$ & $-$ & $-$ & $-$ & $-$ & $-$ & $-$ & $-$ & $-$ & 64.5 & $-$ & $-$ & $-$ \\
SOLO~\cite{solo}         & 7B & $-$ & $-$ & $-$ & 1260 & 34.4 & 64.4 &  $-$ & $-$ & $-$ & $-$ & $-$ & 73.3 & $-$ & $-$ & 61.4 & $-$ & $-$ & $-$ \\
Chameleon-7B\footnotemark[1]~\cite{team2024chameleon} & 7B & 31.1 & 8.3 & 25.4 & 170 & 22.3 & 30.6 & 7 & 17.1 & 3.5 & 16.1 & 4.8 & 47.2 & $-$ & 1.5 & 46.0 & 2.9 & 5.0 & 17.9 \\
EVE-7B~\cite{diao2024EVE}       & 7B & 49.5 & 25.6 & 32.3 & 1483 & 25.2 & 61.3 & 327 & 21.1 & 12.4 & 34.8 & 51.9 & 63.0 & 60.8 & 22.0 & 48.5 & 19.5 & 20.0 & 40.8 \\
EVE-7B (HD)~\cite{diao2024EVE}  & 7B & 52.3 & 25.7 & 32.6 & 1628 & 34.2 & 64.6 & 398 & 26.4 & 16.3 & 38.9 & 56.8 & 64.9 & 62.6 & 53.0 & 61.0 & 59.1 & 25.0 & 54.6 \\
Emu3~\cite{emu3}         & 8B & 58.5 & 37.2 & 31.6 & $-$ & $-$ & \textbf{68.2} & 687 & $-$ & $-$ & $-$ & 64.7 & 89.2 & 60.3 & 76.3 & \textbf{70.0} & 68.6 & \textbf{43.8} & 67.6 \\
\rowcolor{gray!15}
Mono-InternVL-2B & 1.8B & \textbf{65.5} & \textbf{40.1} & \textbf{33.7} & \textbf{1875} & \textbf{45.7} & 67.4 & \textbf{767} & \textbf{34.8} & \textbf{66.3} & \textbf{55.2} & \textbf{72.6} & \textbf{93.6} & 59.5 & \textbf{80.0} & 68.6 & \textbf{73.7} & 43.0 & \textbf{70.1} \vspace{-0.6mm} \\
\bottomrule
\end{tabular}
}
\caption{\textbf{Comparison with existing MLLMs on general MLLM benchmarks and visual question answering benchmarks.} ``\#A-Param'' denotes the number of activated parameters. For MME, we sum the perception and cognition scores.   Avg$_{\text{MM}}$ and Avg$_{\text{QA}}$ denote the  normalized average performance of MLLM benchmarks and VQA benchmarks, respectively. $^\dagger$ InternVL-1.5-2B adopts the same LLM and high-quality data with Mono-InternVL-2B, so we mark it as the modular counterpart. \textbf{Bold} indicates the highest among monolithic MLLMs.}

\label{tab:combined_benchmark}
\end{table*}

\begin{table*}[t]
\centering
\renewcommand{\arraystretch}{1}
\resizebox{0.7\linewidth}{!}{
\begin{tabular}{lccc|ccccc}
\toprule
\multicolumn{1}{l}{\textbf{Model}} &\textbf{ \#A-Param}   & \textbf{Data}  & \textbf{Shots} & \textbf{COCO Caps}  & \textbf{Flickr30k} & \textbf{NoCaps}  & \textbf{VQAv2} \\ \midrule
Flamingo~\cite{VLP:Flamingo}                    & 3B     &  \textgreater  2.1B    & 0     & 73.0  & $-$        & $-$           & 49.2  \\
MM1~\cite{VLM:MM1}                         & 3.5B   & \textgreater  2.3B  & 0     & 73.5  & $-$  & 55.6     & 46.2  \\
Chameleon~\cite{team2024chameleon}                   & 34B   & \textgreater 1.4B   & 2     & 120.2 & 74.7      & $-$        & 66.0  \\
\rowcolor{gray!15} Mono-InternVL-S1.2             & 1.8B& 0.9B&  0     & 87.3   & 72.7          &  54.1      &   $-$           \\
\rowcolor{gray!15} Mono-InternVL-S1.3             & 1.8B &1.1B& 0     & \textcolor{gray}{135.6}      &  77.3         & 116.5       &        \textcolor{gray}{71.1}        \vspace{-0.5mm} \\ 
\bottomrule
\end{tabular}
}
\caption{\textbf{Zero-shot pre-training performance of Mono-InternVL and existing MLLMs.}  ``S1.2'' and ``S1.3'' denote pre-training stages of semantic learning and alignment learning, respectively. Images of COCO have been seen in S1.3, so we mark its performance in gray.}
\vspace{-1em}
\label{few_shot_result}
\end{table*}

\begin{table}[t]
\centering
\renewcommand{\arraystretch}{1.2}
\setlength\tabcolsep{1.5pt}
\resizebox{0.6\linewidth}{!}{
\begin{tabular}{lc|ccccc}
\toprule
\multicolumn{1}{l}{\textbf{Models}} & \multicolumn{1}{l|}{\textbf{\#A-Param}} & \textbf{MMLU} & \textbf{CMMLU} & \textbf{AGIEval} & \textbf{MATH}  \\ \hline
InternLM2-Chat~\cite{cai2024internlm2}       & 1.8B                           & 47.1 &  46.1     & 38.8    &    13.9          \\ 
\hline
EVE~\cite{diao2024EVE}                        & 7B                             &   43.9   &   33.4    &     22.6    &     0.7        \\
Chameleon~\cite{team2024chameleon}                  & 7B                             & 52.1 & -     & -       & 11.5         \\
\rowcolor{gray!15} Mono-InternVL              & 2B                             & 45.1 &   44.0    & 40.9   & 12.3\vspace{-0.6mm}          \\ \bottomrule
\end{tabular} }
\vspace{-.5em}
\caption{\textbf{Comparison of Mono-InternVL and existing monolithic MLLMs on four common NLP tasks. } Except for Chameleon, models are evaluated using OpenCompass toolkit~\cite{opencompass2023}. }
\label{NLP_TABLE}
\end{table}

\subsection{Evaluation Benchmarks}
\vspace{-1mm}
We evaluate Mono-InternVL and existing MLLMs on 16 comprehensive multimodal benchmarks and 4 NLP benchmarks. 
Specifically,  MLLM benchmarks encompass MMBench-EN \textit{test}~\citep{Datasets:MMBench}, MMVet~\citep{Datasets:MM-vet}, MMMU \textit{val}~\citep{Datasets:MMMU}, MME~\citep{Datasets:MME}, MathVista \textit{testmini}~\citep{Datasets:Mathvista}, SEED Image~\citep{Datasets:Seed-bench}, OCRBench~\citep{liu2023ocrbench}, HallusionBench~\citep{Datasets:Hallusionbench}, and CCBench \textit{dev}~\citep{Datasets:MMBench}.
Visual question answering benchmarks include TextVQA \textit{val}~\citep{Datasets:TextVQA}, SQA \textit{test}~\citep{Datasets:ScienceQA}, GQA \textit{test-dev}~\citep{Datasets:GQA}, DocVQA \textit{test}~\citep{Datasets:DocVQA}, AI2D \textit{test}~\citep{Datasets:AI2D}, ChartQA \textit{test}~\citep{Datasets:ChartQA}, and InfographicVQA \textit{test} \citep{mathew2022infographicvqa}.
NLP benchmarks include MMLU~\cite{MMLU}, CMMLU~\cite{CMMLU}, AGIEval~\cite{AGIEVAL} and MATH~\cite{MATH}.
The evaluation metrics follow  existing methods~\citep{VLM:InternVL-1.5,diao2024EVE}.
Some results of Chameleon and EVE are evaluated with VLMEvalKit~\citep{duan2024vlmevalkit} or from the OpenCompass leaderboard~\citep{opencompass2023}.

\subsection{Implementation Details}
Mono-InternVL is implemented based on InternLM2-1.8B~\citep{cai2024internlm2} with newly added visual tokenizer and visual experts.  The visual experts are initialized from  pre-trained MLPs in  InternLM2-1.8B to leverage existing learned representations for improved visual feature extraction, which accounts for 1.2 billion parameters. We adopt a similar dynamic high-resolution strategy from InternVL-1.5~\citep{VLM:InternVL-1.5} to align an optimal resolution for input image, which is then patchfied to visual tokens.  The remaining configurations are kept identical to InternLM2-1.8B. The endogenous visual pre-training  and instruction tuning take approximately 16 days (646\textit{k} iterations) and 1 day (14\textit{k} iterations) on 256 NVIDIA A100 GPUs, respectively.  More  detailed training configurations are  given in Appendix \S\ref{sec:supp_training_detail}.

\subsection{Quantitative Experiments}
\begin{table*}[t]
\centering
\renewcommand{\arraystretch}{1.2}
\setlength\tabcolsep{4.5pt}
\vspace{-1mm}
\resizebox{0.85\linewidth}{!}{
\begin{tabular}{lcc|ccccccc}
\toprule
\multicolumn{1}{l}{\textbf{Model}} & \textbf{\#T-Param} & \textbf{Strategy} & \textbf{MME-P} & \textbf{DocVQA} & \textbf{InfoVQA} &\textbf{SQA-I}& \textbf{GQA} & \textbf{ChartQA} & \textbf{AI2D} \\ 
\midrule
InternLM2                   & 1.8B   & Full   &   753    & 16.1   & 11.6  &  36.7   &  51.4   &   10.8     &  27.7    \\
+ V-Expert                      & 3.0B   & Full   &  948     &  18.6  &  11.9 &  37.7    &  53.0   & 11.1       & 26.6 \\
+ V-Expert                      & 1.2B   & Delta   &    995   &  18.9  & 14.6 & 56.5   &  53.4   &    13.5    & 42.7 \\
\bottomrule
\end{tabular}
}
\caption{\textbf{Ablation of different strategies for visual pre-training.}  All models are pre-trained on 61 million image-text pairs from Laion-2B~\cite{Datasets:Laion-5b} and fine-tuned on instruction data from LLaVA-665\textit{k}.~\cite{VLM:LLaVA-1.5}.  ``Full'' and ``Delta'' denote full tuning and delta tuning, respectively.  ``T-Param'' refers to trainable parameters.} 
\label{ablation_mmoe}
    \vspace{-.5em}
\end{table*}
\begin{figure*}[t]
    \centering
    \includegraphics[width=0.95\linewidth]{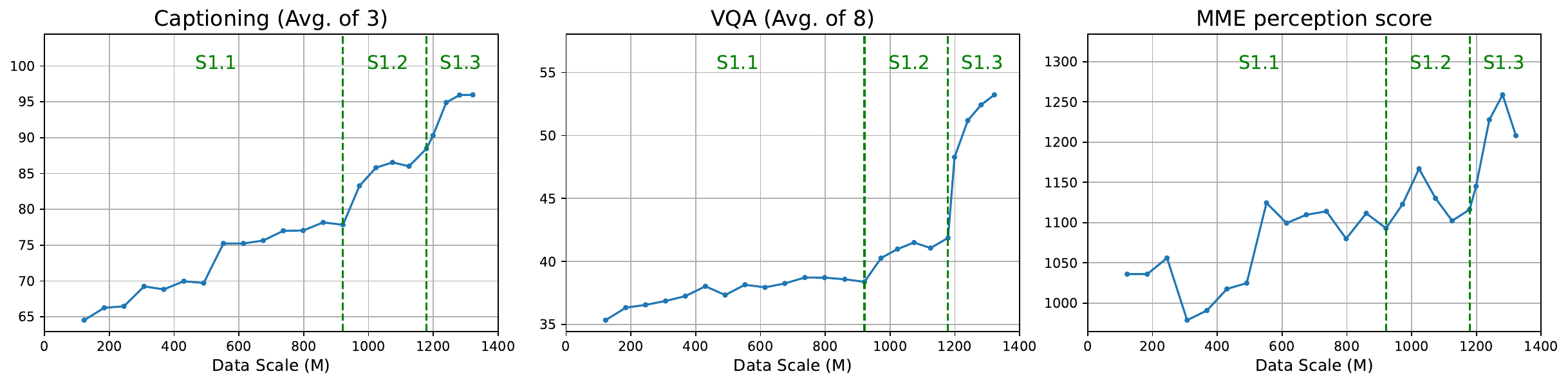}
    \vspace{-1em}
    \caption{\textbf{Ablation studies of EViP with the increase of pre-training data size across three sub-stages: (S1.1) Concept learning; (S1.2) Semantic learning; (S1.3) Alignment learning.} For each data point, we fine-tune the corresponding pre-trained model on the instruction data of LLaVA-665\textit{k} and obtain the downstream performance.  Results of captioning and VQA are averaged from 3 and 8 tasks, respectively. See Appendix \S\ref{sec:supp_full_data_curve}  for complete results.}
    \label{fig:data_curve_avg}
\end{figure*}

\textbf{Comparison with existing MLLMs.} In Tab.~\ref{tab:combined_benchmark}, we compare Mono-InternVL and existing MLLMs on 16 multimodal benchmarks.  The first observation is that  most modular MLLMs outperform existing monolithic MLLMs by significant margins. For example,  the average performance of  InternVL-1.5-2B~\citep{VLM:InternVL-1.5} on 9 MLLM benchmarks greatly exceeds the SoTA monolithic MLLM, \emph{i.e.,} + 15.5\% over EVE-7B (HD)~\citep{diao2024EVE}.  These results strongly suggest  the challenges  in  existing monolithic MLLMs.  In contrast,  Mono-InternVL-2B with a slightly smaller model size can even outperform the  modular baseline, \emph{i.e.,} + 0.8\% against  InternVL-1.5-2B on average.  Notably, Mono-InternVL-2B demonstrates distinct advantages on MathVista and OCRBench, suggesting its seamless text recognition and reasoning capabilities.  Compared to SoTA modular MLLM, \emph{i.e.,} Qwen2VL~\cite{Qwen2vl}, Mono-InternVL-2B  is still comparable in most benchmarks, \emph{e.g.,} MME and MathVista. We believe that better instruction data and stronger LLM will further eliminate this gap. We also observe that Mono-InternVL is still inferior to  InternVL-1.5 on high-resolution benchmarks, \emph{e.g.,} -12.4\% on InfoVQA. This may be because the relatively shallow model depth limits the visual encoding ability of Mono-InternVL, as shown in Fig.~\ref{vis:attention}. 

\footnotetext[1]{Chameleon-7B frequently rejects to perform the task with a response of ``I can't help you with this", thus resulting in poor performance.}

Compared to existing monolithic MLLMs, performance gains of Mono-InternVL become  distinct. For example, compared to  EVE-7B (HD)~\citep{diao2024EVE},   Mono-InternVL achieves up to 15.4\% average gains on   VQA tasks. Note that EVE-7B requires high-quality data for pre-training, so scaling it with more data remains challenging.  Meanwhile, compared to Emu3~\citep{emu3}, the state-of-the-art monolithic MLLM,   Mono-InternVL still demonstrates better results on 9 of 12 benchmarks, while using much fewer parameters. In Tab.~\ref{NLP_TABLE}, we further compare the NLP ability of Mono-InternVL with existing monolithic MLLMs.  From this table, we notice that Mono-InternVL can well preserve its pre-trained NLP ability, which retains similar performance with InternLM2-Chat. However,   monolithic MLLMs like EVE, even with larger parameter size, are still inferior to Mono-InternVL in multiple NLP tasks.  These results further confirm the advantages of Mono-InternVL against existing monolithic MLLMs.

In Tab.~\ref{few_shot_result}, we further compare the pre-training performance of Mono-InternVL and existing MLLMs.  From this table, we can observe that with concept and semantic learning, Mono-InternVL-S1.2 already exceeds existing modular MLLMs, \emph{e.g.,} +13.8 CIDEr over MM1~\citep{VLM:MM1} on COCO Captions, demonstrating that Mono-InternVL-S1.2 is effective in capturing fundamental multimodal relationships.
Compared to monolithic MLLMs, Mono-InternVL also demonstrates superior performance. For instance, even though Chameleon has a much larger model size, it is still inferior to Mono-InternVL-S1.3 by -2.6 CIDEr on Flickr30k~\citep{Datasets:Flickr30k}. These results further confirm  the effectiveness of EViP. 

\begin{table*}[t]
\centering
\renewcommand{\arraystretch}{1.2}
\setlength\tabcolsep{10pt}
\resizebox{0.75\linewidth}{!}{
\begin{tabular}{lccc|ll}
\toprule

\multirow{2}{*}{\textbf{Model}} & \textbf{\#Image} & \textbf{\#Text} & \textbf{\#Total Input} & \multirow{2}{*}{\textbf{TTFT}} & \multirow{2}{*}{\textbf{TPS}} \\
& \textbf{Tokens} & \textbf{Tokens} & \textbf{Tokens} & & \\
\hline
InternVL-1.5-2B  & 768 & 256 & 1024 & 0.24 & 382\\
\rowcolor{gray!15} Mono-InternVL-2B & 768 & 256 & 1024 & 0.09 \textcolor{cvprblue}{\small(-63\%)} & 436 \textcolor{cvprblue}{\small(+14\%)}\\
\hline
InternVL-1.5-2B  & 1792 & 256 & 2048 & 0.45 & 183 \\
\rowcolor{gray!15} Mono-InternVL-2B & 1792 & 256 & 2048 & 0.15 \textcolor{cvprblue}{\small(-67\%)} & 232 \textcolor{cvprblue}{\small(+27\%)} \\
\hline
InternVL-1.5-2B  & 3840 & 256 & 4096 & 1.93 & 52 \\
\rowcolor{gray!15} Mono-InternVL-2B & 3840 & 256 & 4096 & 0.79 \textcolor{cvprblue}{\small(-59\%)} & 68 \textcolor{cvprblue}{\small(+31\%)}\vspace{-0.6mm} \\
\bottomrule
\end{tabular}
}
\caption{\textbf{Inference speed comparison of Mono-InternVL and InternVL-1.5.} Models are deployed on an NVIDIA A100 using LMDeploy with Pytorch backend~\cite{2023lmdeploy}, with a concurrency of 16 and the number of output tokens fixed as 120. ``TTFT'' and ``TPS'' denotes the time to first token in seconds and throughput in tokens per second, respectively.
}
\label{tab:speed}
\end{table*}

 \textbf{Ablation studies.} To validate the design of  Mono-InternVL, we conduct extensive ablation studies in Tab.~\ref{ablation_mmoe} and Fig.~\ref{fig:data_curve_avg}.   Specifically, Tab.~\ref{ablation_mmoe} compares different strategies for visual pre-training.  The first row is the common strategy used in  existing monolithic MLLMs, \emph{i.e.,} full tuning of the LLM,  which yields the worst downstream performance in the table. After employing visual experts (the second row), such a full-tuning strategy becomes more effective, \emph{e.g.,} +1.6\% on GQA.  These comparisons well confirm the sub-optimal  design of the shared architecture for joint vision and language modeling.  Besides, we also observe that the delta tuning strategy greatly benefits the visual pre-training, providing +18.8\% and 16.1\% gains on SQA-I and AI2D, respectively.  Compared to full tuning,  delta tuning can effectively preserve the knowledge of the pre-trained LLM, which is also crucial for   multimodal modeling.
 
Fig.~\ref{fig:data_curve_avg}  further demonstrates the relationship between downstream performance and pre-training data size.  From it we can observe that performance of  Mono-InternVL will gradually reach an upper bound in the concept learning.  Through additional semantic learning and alignment learning, capabilities of Mono-InternVL consistently boost as the data size increases.  It is important to note that that the alignment learning  plays a significant role for VQA and MME, which can provide sufficient task-related knowledge, \emph{e.g.,} OCR knowledge. These results not only demonstrate  the data scalability of Mono-InternVL, but also confirm  the advantages of coarse-to-fine learning in EViP.

\textbf{Comparison of inference efficiency.} In Tab.~\ref{tab:speed}, we compare the inference speed of Mono-InternVL and InternVL-1.5 using the popular deployment library LMDeploy~\citep{2023lmdeploy}. From this table, we can find that due to the elimination of visual encoder, Mono-InternVL demonstrates superior efficiency under different number of input tokens. In particular, the first-token time  is greatly reduced in Mono-InternVL, \emph{e.g.,} up to 67\% against InternVL-1.5.  Benefiting from this, the overall throughput is correspondingly increased by around 31\%. These results greatly validate the efficiency of Mono-InternVL. We note that this is only an initial attempt, and using further optimization techniques may yield better performance.

\subsection{Qualitative Experiments} 
\begin{figure*}[t!]
    \centering
    \includegraphics[width=0.99\linewidth]{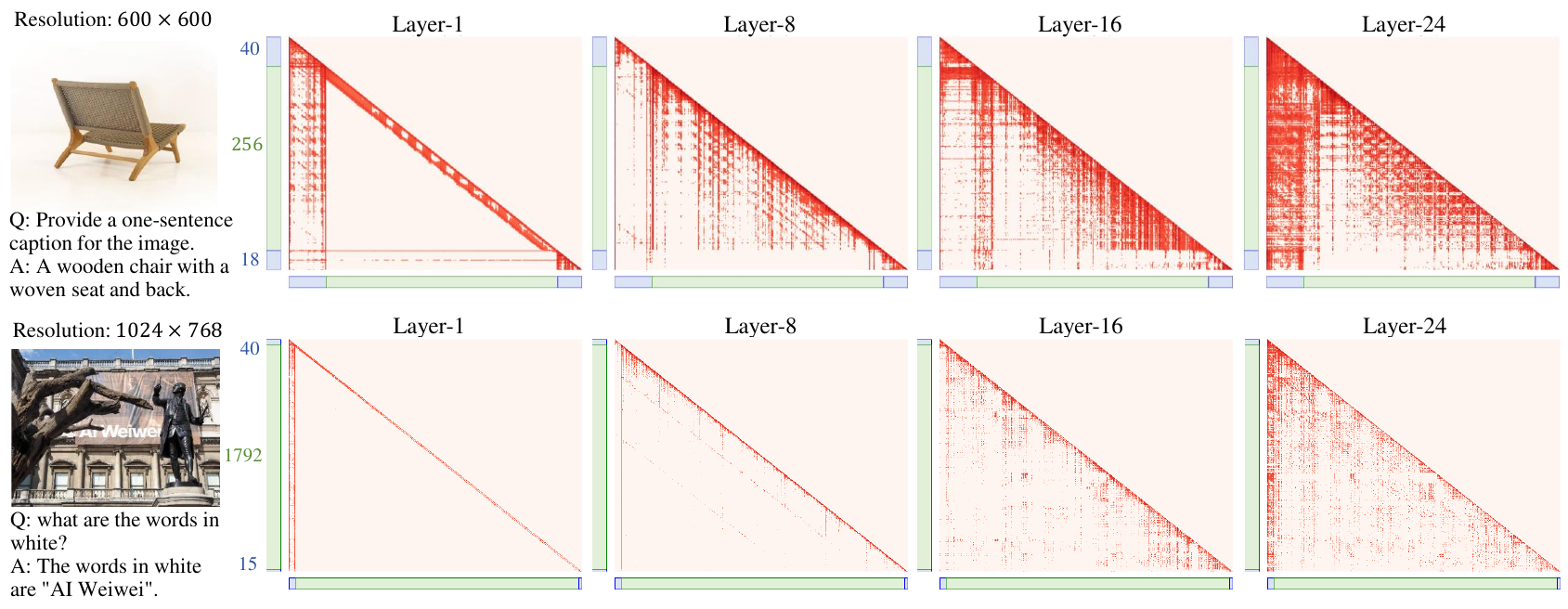}
    \caption{\textbf{Visualization of attention maps in Mono-InternVL.} 
     The first blue segment, green segment and the second green segment  in the axes represent the system prompt tokens (text), image tokens (visual) and user prompt tokens (text), respectively.   The numbers on the left side of  attention maps indicate the number of tokens.
    }
    \label{vis:attention}
\end{figure*}

To gain in-depth insights into Mono-InternVL, we visualize its attention maps of different layers in Fig.~\ref{vis:attention}. From this figure,  we can draw   two noteworthy findings.  Firstly, despite the global connectivity of the Transformer architecture,  locality still exists in the visual encoding of shallow layers.  As shown in Fig.~\ref{vis:attention}, in the first layer, visual tokens only interact with their nearby content, yielding patterns that closely resemble those produced by convolutional neural networks~\citep{CNN:Resnet16}.  Secondly,  modalities are barely interactive at  shallow layers but gradually fused as the layers deepen. As shown in Fig.~\ref{vis:attention},   the attention weights between visual and textual tokens are extremely small in the first layer and  become larger in deeper layers.  We hope these examples will provide  useful hints for the design of monolithic MLLMs.  We provide visualization examples in Appendix \S\ref{sec:supp_visualization}.

\section{Conclusion}
In this paper, we propose Mono-InternVL, a monolithic multimodal large language model (MLLM) that integrates visual encoding and textual decoding  into a single LLM. In Mono-InternVL, a set of visual experts is embedded into the pre-trained LLM  via a mixture-of-experts mechanism. By freezing the  LLM, Mono-InternVL ensures that visual capabilities are optimized without compromising the pre-trained language knowledge.  Based on this structure, an innovative Endogenous Visual Pre-training (EViP) is introduced to realize coarse-to-fine visual learning.   Extensive experiments demonstrate the effectiveness and efficiency of Mono-InternVL compared to existing MLLMs. Our work greatly pushes the boundaries of monolithic MLLMs,  providing new possibilities for the development of MLLMs.

\bibliographystyle{plainnat} 
\bibliography{references}
\clearpage
\appendix
\section{Appendix}
\subsection{More Dataset Details}
\label{sec:supp_full_data_detail}

\begin{table}[htbp]
\small
\renewcommand{\arraystretch}{1.1}
\centering
\renewcommand{\arraystretch}{1.2}
\resizebox{0.7\linewidth}{!}{
\begin{tabular}{l|l}
\toprule
\textbf{Task} & \textbf{Dataset} \\
\hline
Captioning                    & TextCaps (en)~\cite{Datasets:TextCaps}, ShareGPT4V (en\&zh)~\cite{VLM:Sharegpt4v} \\
\rowcolor{gray!15}
                              & VQAv2 (en)~\cite{Datasets:VQAv2}, GQA (en)~\cite{Datasets:GQA}, OKVQA (en)~\cite{Datasets:Ok-vqa},            \\
\rowcolor{gray!15}
\multirow{-2}{*}{General QA}  & VSR (en)~\cite{liu2023vsr}, VisualDialog (en)~\cite{das2017visualdialog}                                       \\
\multirow{-1}{*}{Science}     & AI2D (en)~\cite{Datasets:AI2D}, ScienceQA (en)~\cite{Datasets:ScienceQA}, TQA (en)~\cite{kembhavi2017tqa}      \\
\rowcolor{gray!15}
                              & ChartQA (en)~\cite{Datasets:ChartQA}, MMC-Inst (en)~\cite{liu2023mmcinst}, DVQA (en)~\cite{kafle2018dvqa},     \\
\rowcolor{gray!15}
\multirow{-2}{*}{Chart}       & PlotQA (en)~\cite{methani2020plotqa}, LRV-Instruction (en)~\cite{liu2023lrv-instruction}                       \\

                              & GeoQA+ (en)~\cite{cao2022geoqa_plus}, TabMWP (en)~\cite{lu2022tablemwp}, MathQA (en)~\cite{yu2023mathqa},      \\
\multirow{-2}{*}{Mathematics} & CLEVR-Math/Super (en)~\cite{lindstrom2022clevrmath, li2023superclevr}, Geometry3K (en)~\cite{lu2021geometry3k} \\
\rowcolor{gray!15}
                              & KVQA (en)~\cite{shah2019kvqa}, A-OKVQA (en)~\cite{schwenk2022aokvqa}, ViQuAE (en)~\cite{lerner2022viquae},     \\
\rowcolor{gray!15}
\multirow{-2}{*}{Knowledge}   & Wikipedia (en\&zh)~\cite{he2023wanjuan}                                                                        \\
                              & OCRVQA (en)~\cite{Datasets:OCRVQA}, InfoVQA (en)~\cite{mathew2022infographicvqa}, TextVQA (en)~\cite{Datasets:TextVQA}, \\ 
                              & ArT (en\&zh)~\cite{chng2019art}, COCO-Text (en)~\cite{veit2016cocotext}, CTW (zh)~\cite{yuan2019ctw},          \\
                              & LSVT (zh)~\cite{sun2019lsvt}, RCTW-17 (zh)~\cite{shi2017rctw17}, ReCTs (zh)~\cite{zhang2019rects},             \\
\multirow{-4}{*}{OCR}         & SynthDoG (en\&zh)~\cite{kim2022synthdog}, ST-VQA (en)~\cite{Datasets:STVQA}                                    \\
\rowcolor{gray!15}
Document                      & DocVQA (en)~\cite{Datasets:DocVQA}, Common Crawl PDF (en\&zh)                                                   \\
Grounding                     & RefCOCO/+/g (en)~\cite{Datasets:REFCOCO, Datasets:REFCOCOG}, Visual Genome (en)~\cite{Datasets:VisualGenome}                            \\
\rowcolor{gray!15}
                              & LLaVA-150K (en\&zh)~\cite{VLM:LLaVA}, LVIS-Instruct4V (en)~\cite{wang2023lvisinstruct4v},                   \\
\rowcolor{gray!15}
                              & ALLaVA (en\&zh)~\cite{chen2024allava}, Laion-GPT4V (en)~\cite{laion_gpt4v_dataset},                            \\
\rowcolor{gray!15}
\multirow{-3}{*}{Conversation}&  TextOCR-GPT4V (en)~\cite{textocr_gpt4v_dataset}{}, SVIT (en\&zh)~\cite{Datasets:SVIT}                            \\
                              & OpenHermes2.5 (en)~\cite{OpenHermes2_5}, Alpaca-GPT4 (en)~\cite{taori2023alpaca}, COIG-CQIA (zh)~\cite{bai2024coig},                         \\
\multirow{-2}{*}{Text-only}   & ShareGPT (en\&zh)~\cite{zheng2023vicuna}                                   \\
\bottomrule
    \end{tabular}}
\caption{\textbf{Summary of datasets used in instruction tuning.}}

\label{tab:datasets_finetune}
\end{table}

The datasets used in the instruction fine-tuning stage are listed in Tab.~\ref{tab:datasets_finetune}.
\subsection{More Training Details}
\label{sec:supp_training_detail}

\begin{table}[htbp]
    \centering
        \vspace{-.5em}
    \resizebox{0.8\linewidth}{!}{
    \begin{tabular}{l cccc}
         \toprule
         \multirow{2}{*}{Configuration}     & Concept & Semantic & Alignment & Instruction \\
         & Learning (S1.1) & Learning (S1.2) & Learning (S1.3) & Tuning (S2) \\
         \midrule
                  Maximum numer of patches   & $1,280$ & $1,792$ & $3,328$ & $6,400$ \\
                           LLM sequence length      & $1,425$ & $1,925$ & $4,096$& $8,192$\\
         Use thumbnail            & \multicolumn{4}{c}{\ding{51}} \\
         Optimizer                & \multicolumn{4}{c}{AdamW} \\
         Optimizer hyperparameters & \multicolumn{4}{c}{$\beta_{1}=0.9, \beta_{2}=0.999, eps=1e^{-8}$} \\
         Peak learning rate       & $1e^{-4}$ & $1e^{-4}$ & $5e^{-5}$ & $2e^{-5}$\\
         Learning rate schedule   & constant with warm-up & constant with warm-up & cosine decay & cosine decay\\
         Drop path rate & \multicolumn{4}{c}{$0.1$} \\
         Weight decay             & \multicolumn{4}{c}{$0.01$} \\
         Training steps           & $450$k & $126$k & $70$k & $14$k\\
         Warm-up steps            & $100$ & $100$ & $100$ & $420$ \\
         Global batch size        & $2,048$ & $2,048$ & $2,048$ & $1,024$\\
         Gradient accumulation            & $1$ & $1$ &$1$ & $4$\\
         Numerical precision      & \multicolumn{4}{c}{$\mathtt{bfloat16}$} \\
         \bottomrule
    \end{tabular}
    }
    \caption{\textbf{Hyper-parameters used in the pre-training and instruction tuning of  Mono-InternVL.}}
    \label{tab:hyperparam}
\end{table}

Hyper-parameters used in the training stages are listed in Tab.~\ref{tab:hyperparam}.

\clearpage

\subsection{More Ablation Studies }
\label{sec:supp_full_data_curve}

\begin{figure*}[htbp]
    \centering
    \includegraphics[width=0.8\linewidth]{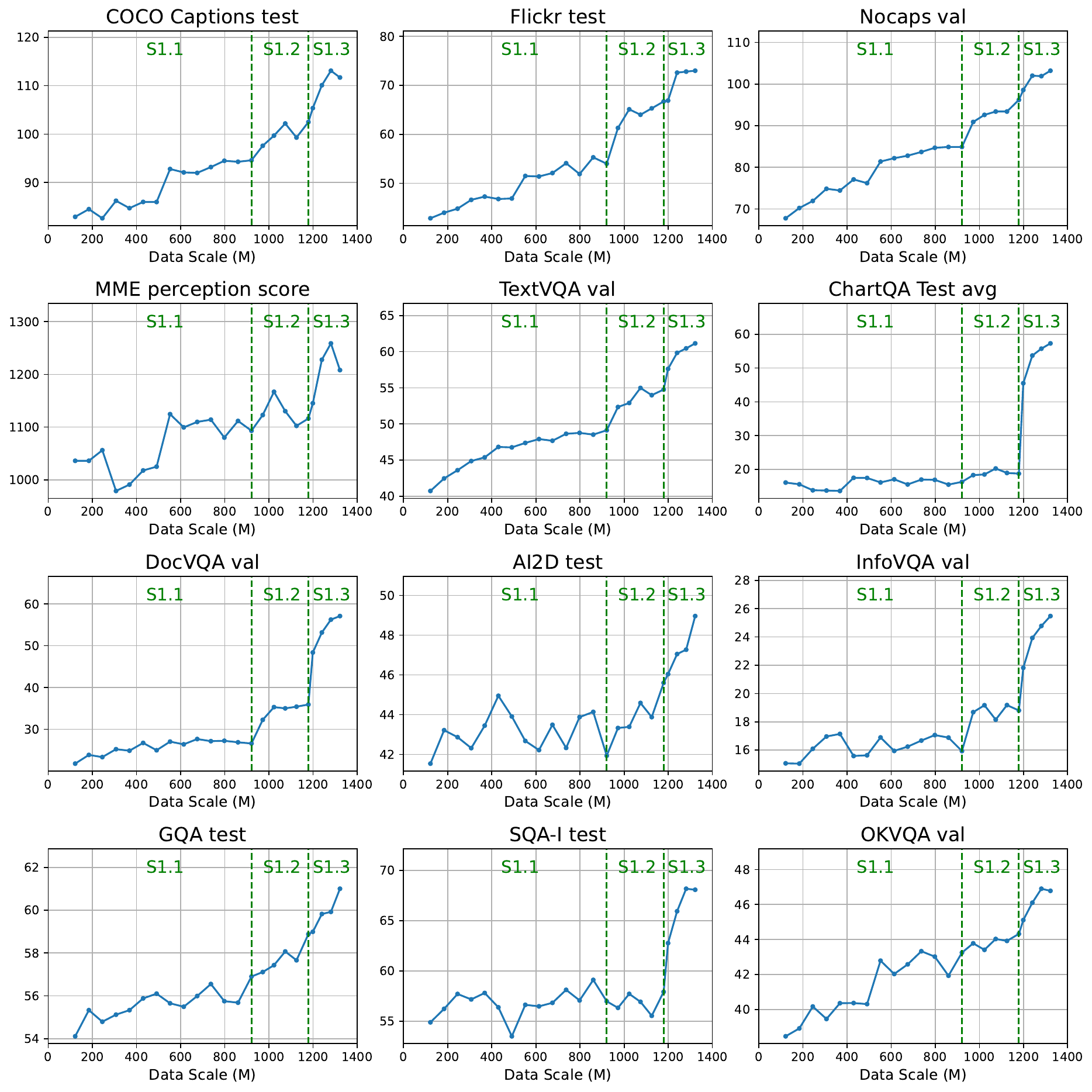}
    \vspace{-1mm}
    \caption{\textbf{Complete results of downstream performance with the increase of pre-training  data size.}}
    \vspace{-1mm}
    \label{fig:data_curve_full}
\end{figure*}

In Fig.~\ref{fig:data_curve_full}, we provide the full results of Fig.~\ref{fig:data_curve_avg} with all the downstream tasks.

In Tab.~\ref{ablation_attention}, we examine the effects of freezing and unfreezing attention layers in alignment learning. We observe that unfreezing attention results in consistent improvements across all metrics, suggesting that it is crucial to optimize the multi-head attentions in this sub-stage for better vision-language alignment.

\label{sec:supp_ablation}

\begin{table}[htbp]
\small
\centering
\renewcommand{\arraystretch}{1.2}
\setlength\tabcolsep{4.5pt}
\vspace{-1mm}
\resizebox{0.7\linewidth}{!}{
\begin{tabular}{lc|ccccccc}
\toprule
\multicolumn{1}{l}{Methods} & \#T-Param   & MME-P & DocVQA & InfoVQA &SQA-I& GQA & ChartQA & AI2D \\ 
\midrule
Freeze attention                   & 1.8B      &   1136   & 39.5   & 19.7  &  56.5   &  59.1   &   27.2     &  44.1    \\
Unfreeze attention                      & 3.0B      &  1153     &  49.3  &  22.7 &  61.8    &  59.9   & 49.5      & 46.4 \\
\bottomrule
\end{tabular}
}
\caption{\textbf{Results of freezing and unfreezing  attention in alignment learning.}  ``T-Param'' refers to trainable parameters. All models are pre-trained on 20 millions of data in alignment learning and fine-tuned on instruction data from LLaVA-665k~\cite{VLM:LLaVA-1.5}. } 
\label{ablation_attention}
\end{table}
\clearpage
 
\subsection{Visualizations}
\label{sec:supp_visualization}
\textbf{Image captioning and OCR}
\begin{tcolorbox}[rounded corners, colback=white, colframe=black, left=5pt, right=5pt, top=5pt, bottom=5pt]
    \begin{minipage}[ft]{0.4\textwidth} 
        \includegraphics[width=\linewidth]{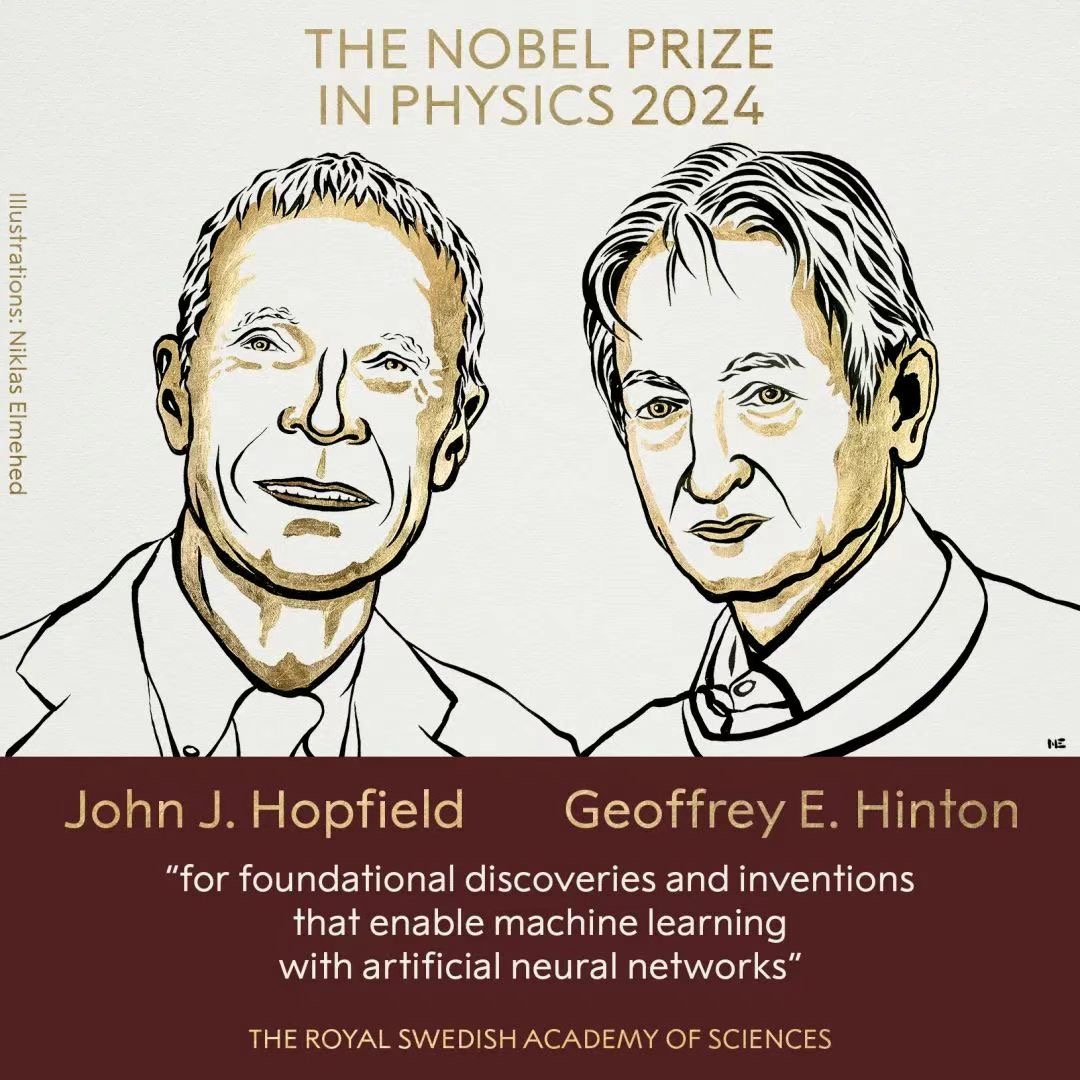}
    \end{minipage} \quad 
    \begin{minipage}[f]{0.56\textwidth} 
    \vspace{2mm}
    \small
    {\color{prompt}\textbf{Prompt:}} What is the content in this picture?  \\

    {\color{prompt}\textbf{Response:}} The image is a graphical representation, primarily informational in nature. It features two individuals depicted in line art style. The background of the image is lightcolored. \\

At the top of the image, there's text that reads ``THE NOBEL PRIZE IN PHYSICS 2024.'' This indicates that the award being discussed in this image pertains to physics and is scheduled for the year 2024. \\

Below this header, there are two names mentioned: John J. Hopfield and Geoffrey E. Hinton. 
Further down, there is a quote attributed to both named individuals: ``for foundational discoveries and inventions that enable machine learning with artificial neural networks.''
This quote suggests that both John J. Hopfield and Geoffrey E. Hinton made significant contributions to the field of artificial intelligence and neural networks, which may have involved pioneering discoveries and innovations that were crucial for developing machine learning technologies. \\

Finally, at the bottom, there's another text stating: ``THE ROYAL SWEDISH ACADEMY OF SCIENCES.''
This indicates that John J. Hopfield and Geoffrey E. Hinton have been recognized by the Royal Swedish Academy of Sciences. \\

Overall, this image appears to serve as a tribute or recognition of the achievements of these
scientists in the field of physics, particularly relating to artificial intelligence and neural networks.
    \end{minipage}
\end{tcolorbox}

\begin{tcolorbox}[rounded corners, colback=white, colframe=black, left=5pt, right=5pt, top=5pt, bottom=5pt]
    \begin{minipage}[ft]{0.4\textwidth} 
        \includegraphics[width=\linewidth]{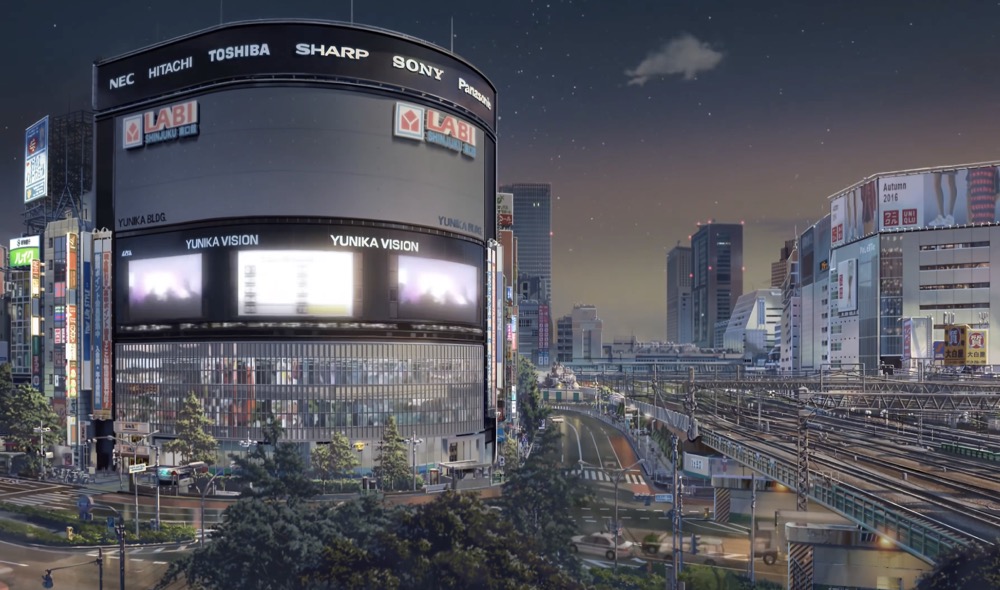}
    \end{minipage} \quad 
    \begin{minipage}[f]{0.56\textwidth} 
    \vspace{2mm}
    \small
    {\color{prompt}\textbf{Prompt:}} Describe this image.  \\

    {\color{prompt}\textbf{Response:}} This image depicts a bustling urban scene at night in an animation style, with a large electronic billboard prominently displaying the words "YUNIKA VISION" in bright white letters against a black background. The billboard is surrounded by several other tall buildings, including those of well-known companies such as HITACHI, TOSHIBA, SHARP, SONY, PANASONIC, and NEC.

In the foreground, there are multiple train tracks that intersect near what appears to be an elevated railway or subway system. Several trains are visible, indicating active transportation amidst the busy cityscape.

The skyline is dotted with various other commercial and residential structures, contributing to the sense of a thriving metropolis. Trees can also be seen around the area, providing some greenery amidst the urban environment. Overall, the image captures the dynamic energy of a modern city at night, with its blend of technology, commerce, and human activity.
    \end{minipage}
\end{tcolorbox}

\clearpage

\textbf{Visual grounding}
\begin{tcolorbox}[rounded corners, colback=white, colframe=black, left=5pt, right=5pt, top=5pt, bottom=5pt]

    \begin{minipage}[ft]{0.19\textwidth} 
        \includegraphics[width=\linewidth]{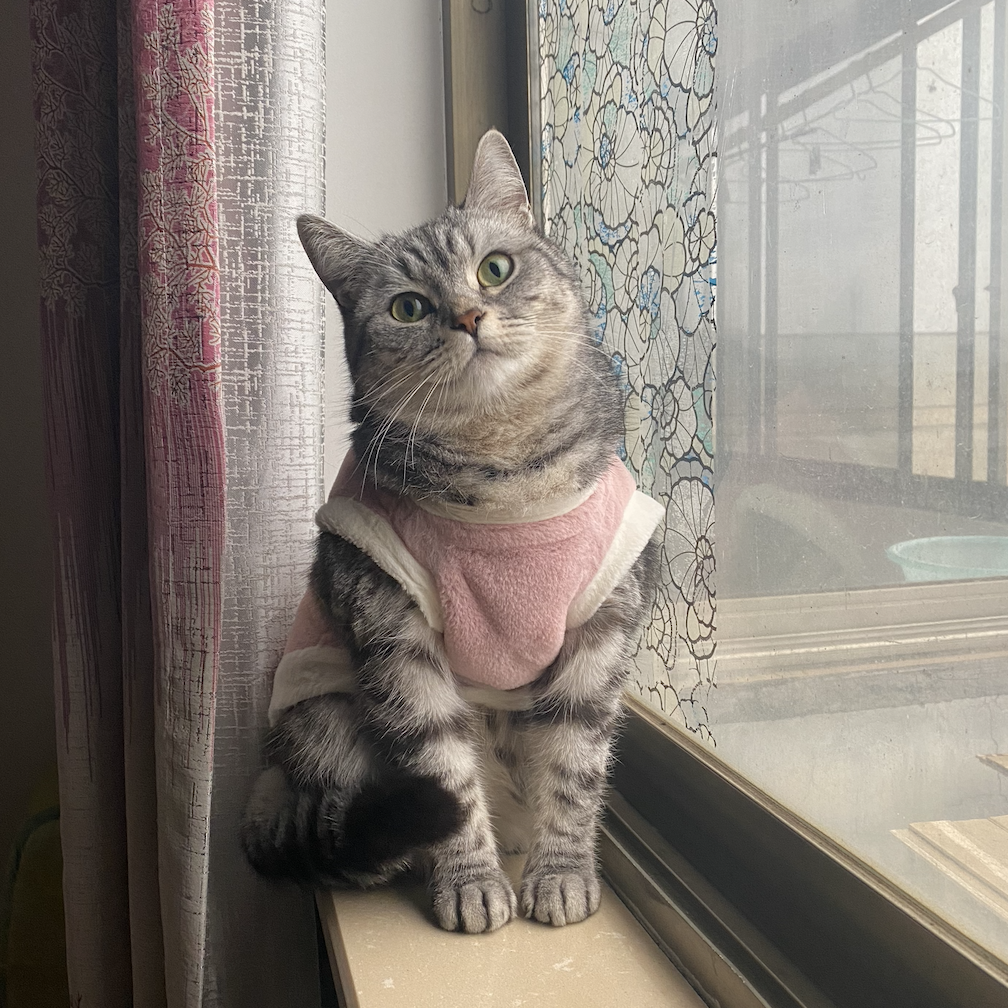} \\
        \includegraphics[width=\linewidth]{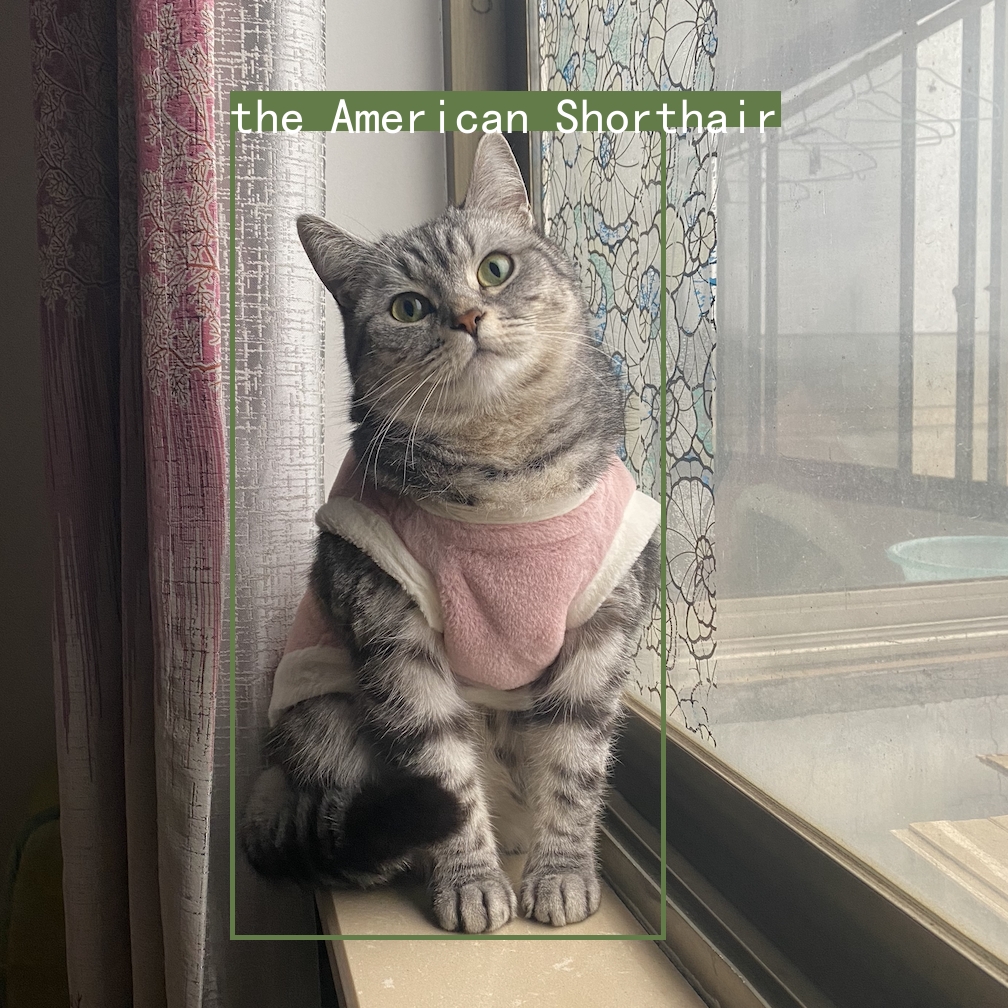} 
    \end{minipage} \quad 
    \begin{minipage}[f]{0.728\textwidth} 
    \vspace{2mm}
    \small
    {\color{prompt}\textbf{Prompt:}} Detect the $<$ref$>$the American Shorthair$<$/ref$>$ in the image with its bounding box.  \\

    {\color{prompt}\textbf{Response:}} $<$ref$>$ the American Shorthair $<$/ref$>$ $<$box$>$ [[229, 125, 660, 932]] $<$/box$>$\\

    {\color{prompt}\textbf{Prompt:}} Please describe the detected object. \\

    {\color{prompt}\textbf{Response:}} The object in the image is a cat. The cat has gray fur with black stripes and is wearing a pink sweater. It appears to be sitting on a windowsill, looking up attentively towards the camera.\\
    \end{minipage}
\end{tcolorbox}

\begin{tcolorbox}[rounded corners, colback=white, colframe=black, left=5pt, right=5pt, top=5pt, bottom=5pt]

    \begin{minipage}[ft]{0.28\textwidth} 
        \includegraphics[width=\linewidth]{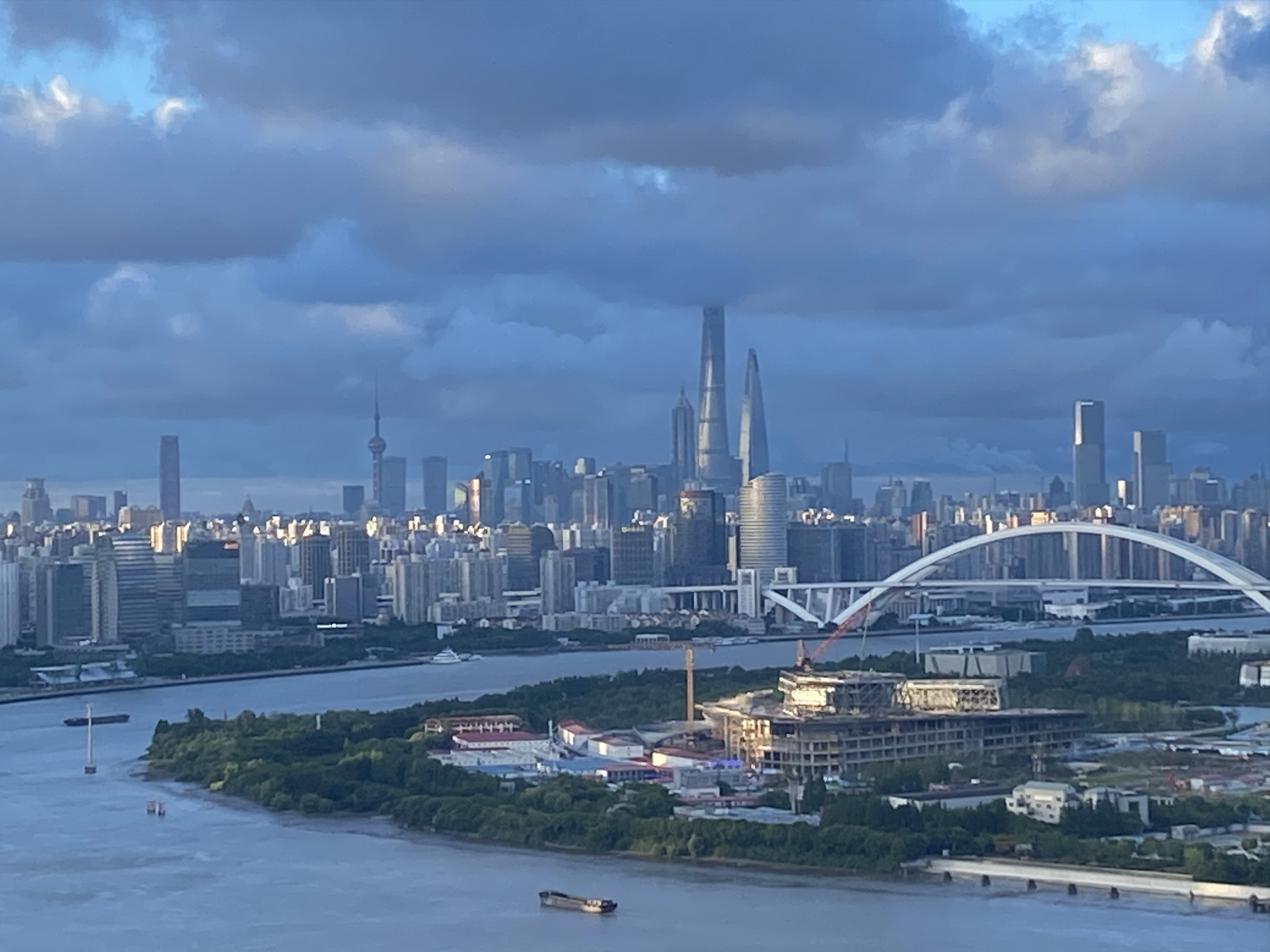} \\
        \includegraphics[width=\linewidth]{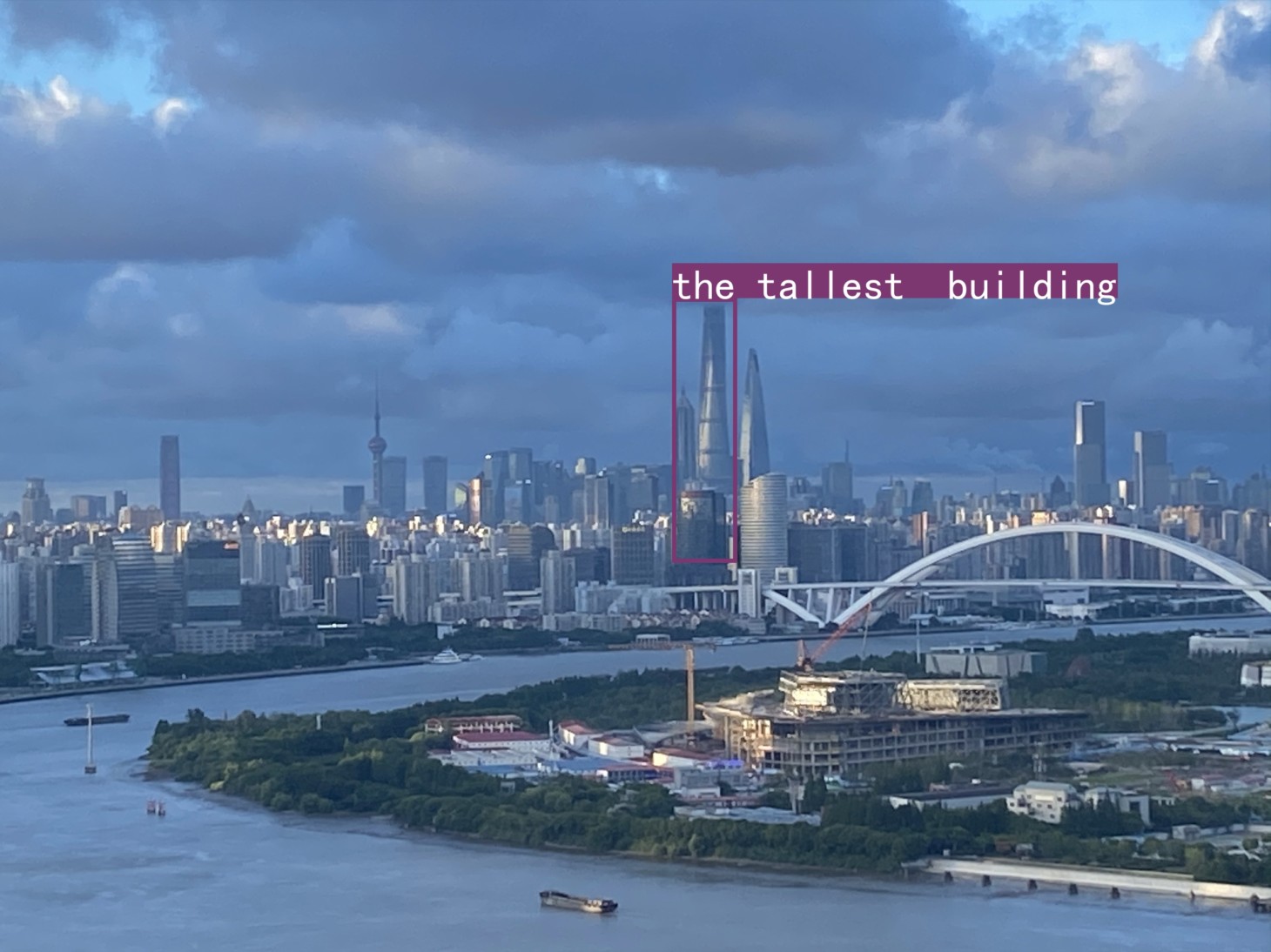} 
    \end{minipage} \quad 
    \begin{minipage}[f]{0.68\textwidth} 
    \vspace{2mm}
    \small
    {\color{prompt}\textbf{Prompt:}} Detect the $<$ref$>$the tallest building$<$/ref$>$ in the image with its bounding box.\\

    {\color{prompt}\textbf{Response:}} $<$ref$>$ the tallest building $<$/ref$>$ $<$box$>$ [[529, 313, 580, 591]] $<$/box$>$\\

    {\color{prompt}\textbf{Prompt:}} Please infer where this place is. \\

    {\color{prompt}\textbf{Response:}} This image depicts a stunning view of the skyline of Shanghai, China. The prominent features in the skyline include several high-rise buildings with unique structures such as the Oriental Pearl Tower and the Shanghai World Financial Center. There is also a distinctive arch bridge spanning across a river, which appears to be part of Shanghai's famous Huangpu River waterfront area known for its historic sites and modern development.\\
    \end{minipage}
\end{tcolorbox}


\textbf{VQA}
\begin{tcolorbox}[rounded corners, colback=white, colframe=black, left=5pt, right=5pt, top=5pt, bottom=5pt]
    \begin{minipage}[ft]{0.28\textwidth} 
        \includegraphics[width=\linewidth]{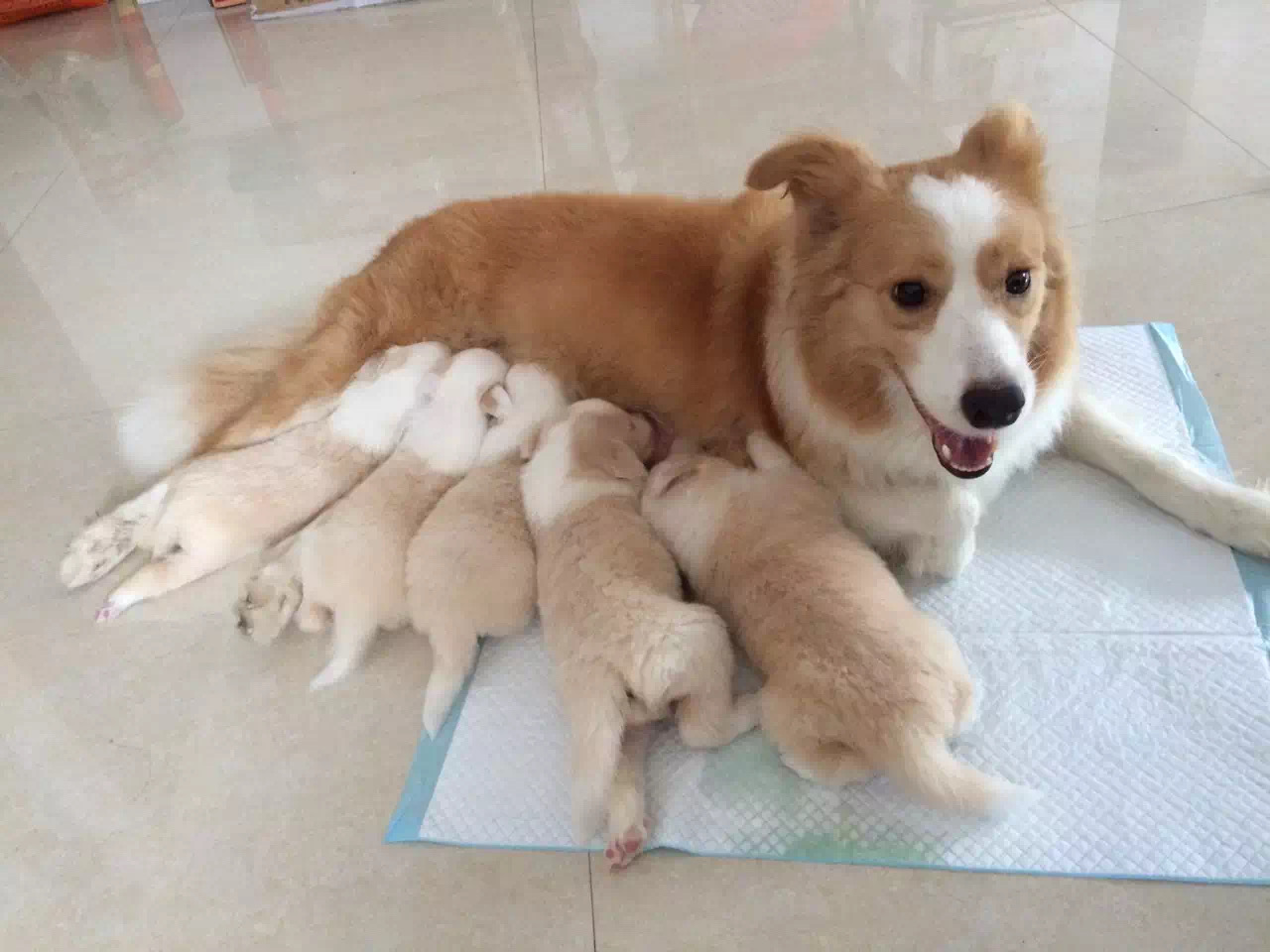}
    \end{minipage} \quad 
    \begin{minipage}[f]{0.68\textwidth} 
    \vspace{2mm}
    \small
    {\color{prompt}\textbf{Prompt:}} Describe the image.  \\

    {\color{prompt}\textbf{Response:}} The image features a large adult dog lying on its side, likely the mother of several puppies. The adult dog has brown and white fur with distinct markings on its face. She is surrounded by several small puppies, all of which are nursing from her body or resting on a light blue towel that covers part of the floor. The setting appears to be indoors, possibly in a home environment given the tiled flooring and the soft, comfortable-looking towel. The atmosphere seems calm and nurturing as the dogs interact closely together.\\

    {\color{prompt}\textbf{Prompt:}} Please tell me the exact number of golden retrievers in the picture. \\

    {\color{prompt}\textbf{Response:}} There are a total of six golden retrievers in the picture.\\
    \end{minipage}
\end{tcolorbox}

\clearpage

\textbf{Chinese OCR}
\begin{tcolorbox}[rounded corners, colback=white, colframe=black, left=5pt, right=5pt, top=5pt, bottom=5pt]
    \begin{minipage}[ft]{0.35\textwidth} 
        \includegraphics[width=\linewidth]{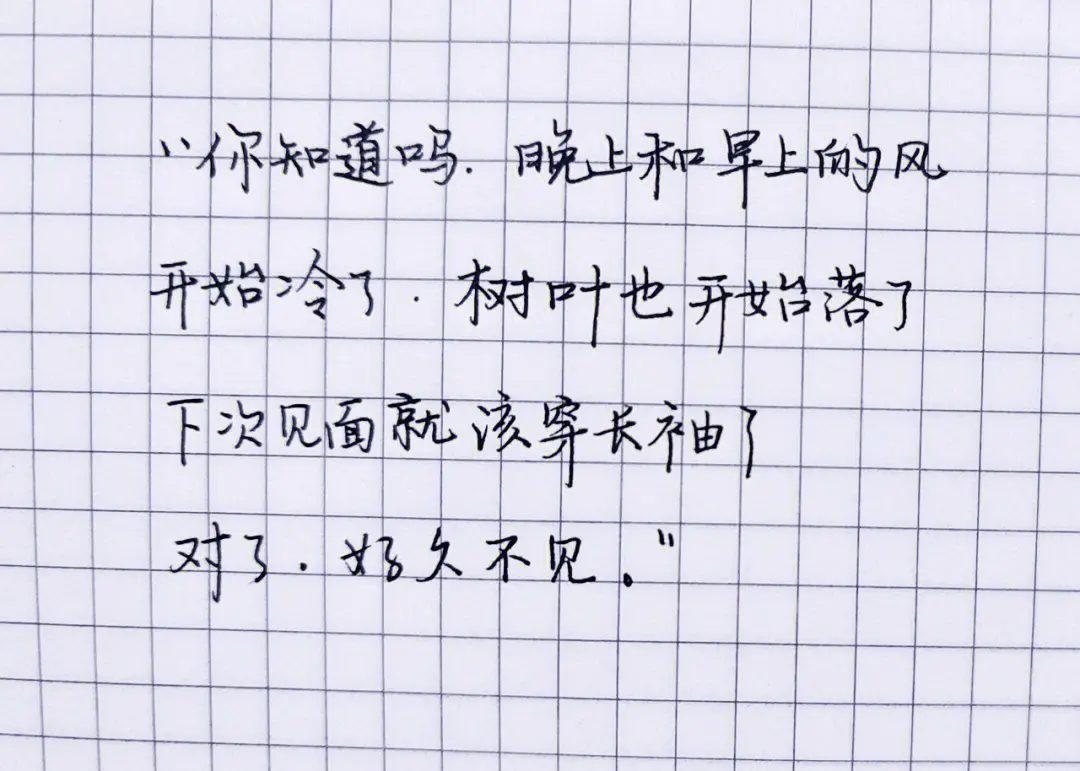}
    \end{minipage} \quad 
    \begin{minipage}[f]{0.56\textwidth} 
    \vspace{2mm}
    \begin{CJK*}{UTF8}{gbsn}
    \small
    
    {\color{prompt}\textbf{Prompt:}} 识别图中文字。  \\

    {\color{prompt}\textbf{Response:}} 图中文字如下：

“你知道吗。晚上和早上的风 开始冷了，树叶也开始落了 下次见面就该穿长袖了 对了。好久不见。”
    \end{CJK*}
    \end{minipage}
\end{tcolorbox}

\textbf{Code generation}
\begin{tcolorbox}[rounded corners, colback=white, colframe=black, left=5pt, right=5pt, top=5pt, bottom=5pt]
    \begin{minipage}[ft]{0.98\textwidth} 
    \centering
        \includegraphics[width=0.6\linewidth]{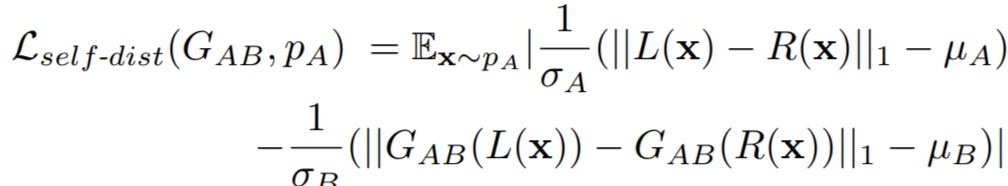}    \end{minipage} 
    \begin{minipage}[f]{0.98\textwidth} 
    \vspace{2mm}
    \small
    {\color{prompt}\textbf{Prompt:}} Convert the formula in the figure to latex format.  \\

    {\color{prompt}\textbf{Response:}} Certainly! Here is the formula from the figure in LaTeX format:\\
    {\scriptsize
    \begin{verbatim}
    \begin{equation}
    \begin{aligned}
        \mathcal{L}_{self-dist}(G_{AB}, p_A) = \mathbb{E}_{\textbf{x}\sim p_A}|
        \frac{1}{\sigma_A}(|| L(\textbf{x}) - R(\textbf{x}) ||_1 - \mu_A)\\
        - \frac{1}{\sigma_B}(||G_{AB}(L(\textbf{x})) - G_{AB}(R(\textbf{x}))||_1 - \mu_B)|
    \end{aligned}
    \end{equation}
    \end{verbatim}
    }

    \end{minipage}
\end{tcolorbox}

\textbf{Document understanding}
\begin{tcolorbox}[rounded corners, colback=white, colframe=black, left=5pt, right=5pt, top=5pt, bottom=5pt]
    \begin{minipage}[ft]{0.35\textwidth} 
        \includegraphics[width=\linewidth]{assets/doc.png}
    \end{minipage} \quad 
    \begin{minipage}[f]{0.56\textwidth} 
    \vspace{2mm}
 
    \small
    {\color{prompt}\textbf{Prompt:}} What is the age of subject B.L.?\\
    
    {\color{prompt}\textbf{Response:}} B.L. is 5 years old at the time of study on the effect of milk protein dilution.\\
    
    {\color{prompt}\textbf{Prompt:}} Whose initial weight is the largest? What is it?  \\

    {\color{prompt}\textbf{Response:}} The largest initial weight in the table is for P.K. with 61.5 pounds (lb.).\\

    {\color{prompt}\textbf{Prompt:}} How many children participated in the experiment in total?\\
    
    {\color{prompt}\textbf{Response:}} There are a total of 14 children who participated in the experiment.
 
    \end{minipage}
\end{tcolorbox}

\clearpage

\textbf{Math}
\begin{tcolorbox}[rounded corners, colback=white, colframe=black, left=5pt, right=5pt, top=5pt, bottom=5pt]
    \begin{minipage}[ft]{0.98\textwidth} 
    \centering
        \includegraphics[width=0.35\linewidth]{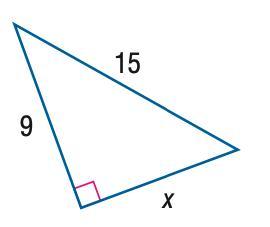}    \end{minipage} 
    \begin{minipage}[f]{0.98\textwidth} 
    \vspace{2mm}

    \small
    {\color{prompt}\textbf{Prompt:}} The figure shows a right-angled triangle. What is the value of $x$?  \\

    {\color{prompt}\textbf{Response:}} To find the value of $x$, we can use the Pythagorean theorem, which states:
$a^2+b^2=c^2$
where c is the hypotenuse (the side opposite the right angle), and a and b are the other two sides.\\
In this triangle:\\
The length of the hypotenuse ($c$) is 15.\\
One leg ($a$) is 9.\\
The other leg ($b$) is $x$.\\
Using the Pythagorean theorem:\\
$9^2+x^2=15^2$\\
$81+x^2=225$\\
Now, solve for $x$:\\
$x^2=225-81$\\
$x^2=144$\\
$x=\sqrt{144}$\\
$x=12$\\
So, the value of $x$ is 12.

    \end{minipage}
\end{tcolorbox}

\end{document}